\documentclass[letterpaper]{article} 
\usepackage[draft]{aaai2026}  
\usepackage{times}  
\usepackage{helvet}  
\usepackage{courier}  
\usepackage[hyphens]{url}  
\usepackage{graphicx} 
\usepackage{natbib}  
\usepackage{caption} 
\usepackage{algorithm}
\usepackage{algorithmic}
\usepackage{amsmath}
\usepackage{amssymb}
\usepackage{multirow}
\usepackage{booktabs}
\usepackage[colorlinks]{hyperref}

\usepackage{newfloat}
\usepackage{listings}
\DeclareCaptionStyle{ruled}{labelfont=normalfont,labelsep=colon,strut=off} 
\floatstyle{ruled}
\newfloat{listing}{tb}{lst}{}
\floatname{listing}{Listing}
\title{Ultra-High-Definition Reference-Based Landmark Image \\
Super-Resolution with Generative Diffusion Prior}
\author{
    Zhenning Shi\textsuperscript{\rm 1, \rm 2}, 
    Zizheng Yan\textsuperscript{\rm 2}, 
    Yuhang Yu\textsuperscript{\rm 2}, 
    Clara Xue\textsuperscript{\rm 2},
    Jingyu Zhuang\textsuperscript{\rm 2},\\
    Qi Zhang\textsuperscript{\rm 2},
    Jinwei Chen\textsuperscript{\rm 2},
    Tao Li\textsuperscript{\rm 1, \rm 3}\thanks{they are corresponding authors}, 
    Qingnan Fan\textsuperscript{\rm 2}\footnotemark[1]
}
\affiliations{
    College of Computer Science, Nankai University\textsuperscript{\rm 1}\\
    Vivo Mobile Communication Co. Ltd\textsuperscript{\rm 2}\\
    Key Laboratory of Data and Intelligent System Security Ministry of Education, China\textsuperscript{\rm 3}\\
    litao@nankai.edu.cn, fqnchina@gmail.com
}

\usepackage{bibentry}

\begin{document}

\maketitle

\begin{abstract}

Reference-based Image Super-Resolution (RefSR) aims to restore a low-resolution (LR) image by utilizing the semantic and texture information from an additional reference high-resolution (reference HR) image. 
Existing diffusion-based RefSR methods are typically built upon ControlNet, which struggles to effectively align the information between the LR image and the reference HR image. 
Moreover, current RefSR datasets suffer from limited resolution and poor image quality, resulting in the reference images lacking sufficient fine-grained details to support high-quality restoration. 
To overcome the limitations above, we propose \emph{TriFlowSR}, a novel framework that explicitly achieves pattern matching between the LR image and the reference HR image. 
Meanwhile, we introduce \emph{Landmark-4K}, the first RefSR dataset for Ultra-High-Definition (UHD) landmark scenarios. 
Considering the UHD scenarios with real-world degradation, in TriFlowSR, we design a Reference Matching Strategy to effectively match the LR image with the reference HR image.  
Experimental results show that our approach can better utilize the semantic and texture information of the reference HR image compared to previous methods. 
To the best of our knowledge, we propose the first diffusion-based RefSR pipeline for ultra-high definition landmark scenarios under real-world degradation. 
Our code and model will be available at \href{https://github.com/nkicsl/TriFlowSR}{https://github.com/nkicsl/TriFlowSR}.

\end{abstract}


\section{Introduction}

Single Image Super-Resolution (SISR) \cite{dong2014learning, kim2016accurate, kim2016deeply, lai2017deep} aims to reconstruct high-resolution (HR) details from a given low-resolution (LR) image. 
Recently, diffusion-based models \cite{ho2020denoising, song2020denoising, song2020score, liu2022flow} have emerged as 
powerful frameworks in generating high-fidelity data. 
Due to their strong generative priors, diffusion-based models have been widely adopted in SISR tasks \cite{wang2024exploiting, yu2024scaling, dong2025tsd}, offering significant improvements in perceptual quality and detail reconstruction.
However, due to the inherently ill-posed nature of SISR, the reconstructed images often suffer from visual artifacts and 
may be inconsistent with the original scene, 
particularly when faced with heavily degraded images. 

\begin{figure}[t]
\centering
\includegraphics[width=0.95\columnwidth]{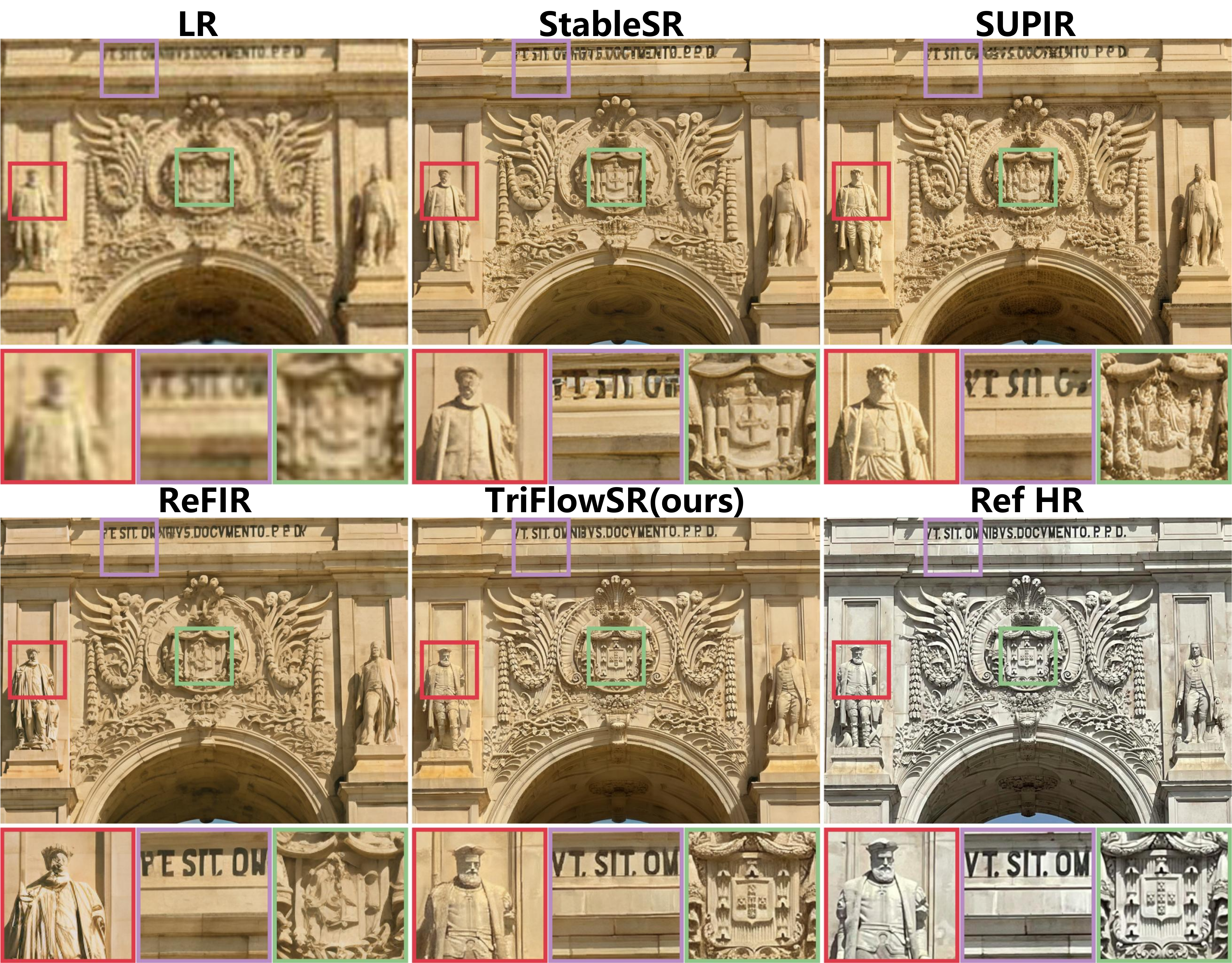} 
\caption{
We propose TriFlow to enable diffusion-based RefSR in UHD landmark scenarios under the real-world degradation. 
Through explicit pattern matching between the LR image and the reference HR image, TriFlow fully leverages the semantic and texture information contained in the reference HR image. 
Please zoom in for a better view.
}
\label{fig: topshow}
\end{figure}

To address the limitations of SISR, Reference-based Super-Resolution (RefSR) has been proposed. 
%
%
Unlike SISR, which relies solely on a single LR image, RefSR introduces a reference HR image to leverage additional structural and texture information, thereby generating more realistic and detailed HR outputs.
However, most existing RefSR approaches \cite{yang2020learning, jiang2021robust, cao2022reference} are built upon relatively simple CNN architectures and 
with idealized degradation assumptions (e.g. bicubic downsampling),
which significantly limits their performance under real-world degradations.

To resolve this problem, recent studies \cite{sun2024coser, guo2024refir} have explored integrating diffusion-based models into the RefSR pipeline. These methods aim to combine the strong generation capabilities of diffusion models with the contextual guidance provided by reference images. Nevertheless, they introduce the reference information via mechanisms like ControlNet \cite{zhang2023adding}, without performing explicit pattern matching between the LR input and the reference HR image. As a result, they still suffer from generative artifacts. In addition, existing RefSR datasets \cite{zhang2019image, jiang2021robust} are limited in both resolution and image quality~(\textit{e.g.}, CUFED5 has an average resolution of 418 $\times$ 418) , leading to reference HR images that lack sufficient fine-grained detail. This is in conflict with the goal of RefSR: (1) RefSR aims to utilize high-quality reference HR images to enhance the restoration results. If the reference HR image itself is of low resolution and poor quality, the restored image quality will not be high either. (2) Images captured by modern smartphones are predominantly Ultra-High-Definition (UHD) images, and low-resolution datasets do not match the actual scenarios. UHD images, with their richer structural and textural details, are inherently better suited for RefSR, enabling more accurate reconstructions that align with current imaging standards and user expectations. Deploying RefSR methods using UHD references could potentially unlock more accurate reconstructions, better aligning with contemporary imaging standards and user expectations.

To overcome these challenges, we propose the \textbf{TriFlowSR} framework. Through the Patch-Ref Attention mechanism, we can explicitly achieve pattern matching between the LR image and the reference HR image, and realize RefSR in real-world degradation scenarios. Meanwhile, we introduce the \textbf{Landmark-4K} dataset, which is the first RefSR dataset in the UHD scenario, consisting of 185 high-quality landmark images covering 49 landmark categories worldwide. Furthermore, to enable the RefSR pipeline in UHD scenarios with real-world degradation, we design the \textbf{Reference Matching Strategy} to effectively match the LR image with the reference HR image.

Our contributions can be summarized as follows:
\begin{itemize}
  \item To the best of our knowledge, we propose the first diffusion-based reference-based super-resolution pipeline for ultra-high definition landmark scenarios under real-world degradation.
  \item We propose the TriFlowSR framework, which can explicitly achieve pattern matching between the LR image and the reference HR image, thereby effectively utilizing the semantic and texture information of the reference HR image.
  \item We propose the first RefSR dataset for UHD scenarios, Landmark-4K. Additionally, we design a Reference Matching Strategy to effectively match the LR image with the reference HR image.
\end{itemize}

\section{Related Works}
\subsection{Single Image Super-Resolution.}
Single Image Super-Resolution (SISR) aims to recover the HR image with only a single LR image as input. SRCNN \cite{dong2014learning} utilized a deep learning-based convolutional neural network for Super-Resolution tasks. Various CNN-based methods have been proposed \cite{kim2016accurate, kim2016deeply, lai2017deep, lim2017enhanced, zhang2018image} to improve the accuracy of reconstructed images. However, CNN-based methods tend to produce overly smooth results and lack high-frequency details. To generate more perceptually realistic images, GANs \cite{goodfellow2020generative, ledig2017photo, zhang2021designing, wang2021real} have been introduced into Super-Resolution tasks. 
While generating more perceptually realistic details, the training of GANs is often unstable and suffers from unnatural visual artifacts. Recently, diffusion-based models \cite{ho2020denoising, song2020denoising, song2020score, liu2022flow} have emerged as powerful and effective conditional generative models, demonstrating remarkable success in synthesizing high-fidelity data and being widely applied to super-resolution tasks \cite{wu2024seesr, wu2024one}. 
StableSR \cite{wang2024exploiting} fine-tuned a time-aware encoder to incorporate low-resolution (LR) image information. SUPIR \cite{yu2024scaling} improved generation by integrating large diffusion backbones with high capacity adapters. TSD-SR \cite{dong2025tsd} distilled a multistep SD3 model \cite{esser2024scaling} into a single-step model by distilling the target score. Although these methods can effectively generate high-fidelity images, their outputs often exhibit noticeable generative artifacts in scenarios with strong degradation, resulting in hallucinations.

\begin{figure*}[ht]
\centering
\includegraphics[width=0.85\textwidth]{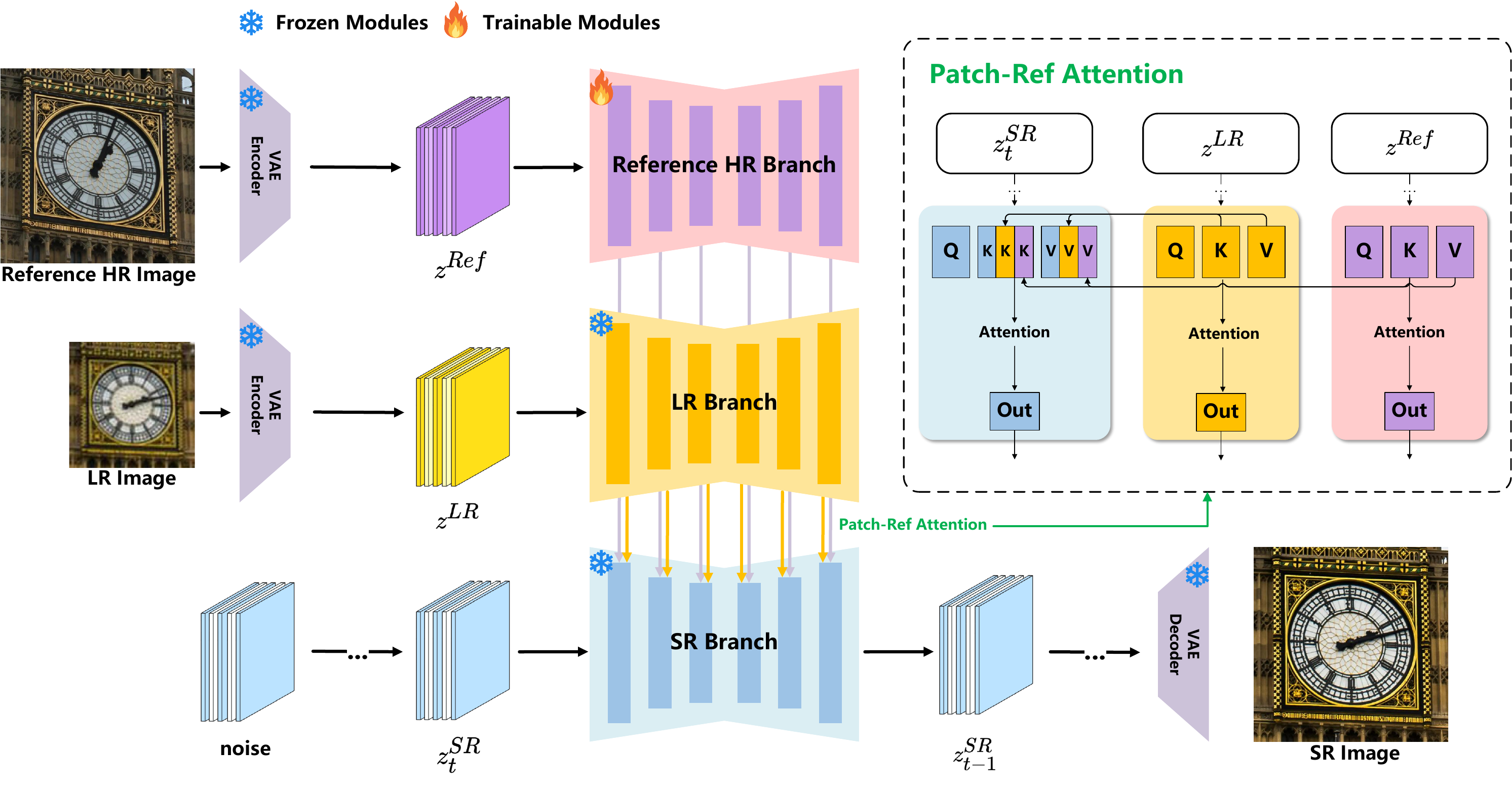} 
    \caption{The framework of our proposed TriFlowSR. TriFlowSR comprises three branches: the SR branch, the LR branch, and the Reference HR branch. During the RefSR training process, the SR and LR branches are frozen, while the LR image and the reference HR image process pattern matching via the Patch-Ref Attention. This enables the transfer of semantic and texture information from the reference HR image.}
\label{fig: TriFlowSR}
\end{figure*}
\subsection{Reference-based Image Super-Resolution.}
Reference-based Image Super-Resolution (RefSR) is dedicated to leveraging the semantic and texture information of an additional reference HR image to guide the restoration of the LR image to the HR image. CrossNet \cite{zheng2018crossnet} aligned the reference and LR images by the flow estimation. 
Later, some works introduced techniques such as multi-scale \cite{zhang2019image}, cross-scale \cite{yang2020learning}, deformable convolution \cite{shim2020robust, dai2017deformable} and coarse-to-fine patch matching \cite{lu2021masa}. 
C2-Matching \cite{jiang2021robust} obtained more accurate pre-offsets of reference features to LR features through a teacher-student correlation distillation.
Based on C2-Matching, some works have explored multi-scale \cite{xia2022coarse}, multi-reference \cite{zhang2023lmr}, and decoupling \cite{huang2022task}. DATSR \cite{cao2022reference} integratesd deformable convolution with the Swin Transformer \cite{liu2021swin}. However, most of these methods are simple CNN models, and they assume simple and known degradations (bicubic), which limits their effectiveness when dealing with complex and unknown degradations in the real world. Recent works have attempted to combine RefSR with diffusion-based models, aiming to retain high-fidelity generation capability while effectively utilize the information from the reference HR image. CoSeR \cite{sun2024coser} combined the information from the LR image and the reference HR image by inputting them into two separate ControlNets. ReFIR \cite{guo2024refir} balanced the reference HR features and the super-resolution (SR) image features through Spatial Adaptive Gating. However, these methods introduce information through ControlNet without performing explicit pattern matching between the LR features and the reference HR features, which still results in generative artifacts.

\section{Methodology}

\subsection{TriFlowSR}
In the RefSR task, we are given two inputs: the LR image $I^{LR}$ and the reference HR image $I^{Ref}$. Our goal is to generate the super-resolution image $I^{SR}$ that closely approximates the distribution of the ground truth HR image $I^{HR}$. Previous diffusion-based RefSR methods introduce the information from the reference HR image into the backbone network through the ControlNet structure, without explicitly performing pattern matching between LR image and the reference HR image. As a result, they struggle to effectively transfer the semantic and texture information from the reference HR image, which will easily result in generative artifacts. To address this issue, we propose the \textbf{TriFlowSR} architecture, which consists of three branches: the Super-Resolution (SR) branch, the Low-Resolution (LR) branch, and the Reference High-Resolution (HR) branch. The training is completed in two stages.

The SR branch is a pre-trained text-to-image (T2I) diffusion model \cite{rombach2022high, esser2024scaling}, which functions by sampling $\epsilon$ from a Gaussian distribution in the latent space to generate high-fidelity images. This branch remains frozen throughout all stages, preserving the diffusion priors and attention mechanisms of the existing pre-trained model. In the first stage, we pre-train the LR branch on the SISR datasets, so that the model has a basic SISR super-resolution capability. The structure of the LR branch is identical to the SR branch. Through the VAE Encoder \cite{van2017neural, esser2021taming}, the LR image $I^{LR}$ is mapped to $z^{LR}=Encoder(I^{LR})$ in the latent space, which transmits the information of the LR image to the SR branch. Let the parameters of the SR branch and the LR branch be $\theta$ and $\Theta$, we have $I^{SISR}=Decoder(\mathcal{F}(\epsilon ,z^{LR};\theta,\Theta))$.

In the second stage, the LR branch and the SR branch are frozen, and training is restricted to the Reference HR branch alone. Similar to the LR branch, the Reference HR branch has the same structure as the SR branch. We have $z^{Ref}=Encoder(I^{Ref})$. To effectively transfer the semantic and textural features of the reference HR image to the backbone network, inspired by \citet{hu2024animate}, we utilize a cross-attention mechanism to convey the information. We propose the \textbf{Patch-Ref Attention}, where the LR image features and the reference HR image features are integrated into the main branch as follows:
\begin{equation}
\begin{aligned}
\label{eq: PatchRefAttn}
PatchRefAttn = &softmax(\frac{Q^{SR}[K^{SR},K^{LR},K^{Ref}]^T}{\sqrt{d_k}})
\\& \cdot [V^{SR},V^{LR},V^{Ref}],
\end{aligned}
\end{equation}
where $Q$, $K$ and $V$ are the query, key and value from the attention layer of
the branches. $[\cdot, \cdot]$ denotes the concatenation operation, and $d_k$ represents the channel dimension. Patch-Ref Attention enables explicit feature matching between the LR image and Reference image at the patch scale. The resulting attention score matrix is then used to selectively retain beneficial reference features while suppressing detrimental features. This interaction enables the model to adaptively select and transfer textures and structural information from the reference HR image rather than relying solely on coarse semantic alignment. Denoting the parameters of the Reference HR branch as $\psi$, we have $I^{RefSR}=Decoder(\mathcal{F}(\epsilon ,z^{LR}, z^{Ref};\theta,\Theta, \psi))$. Following the Rectified Flow\cite{liu2022flow}, our diffusion loss at the second stage can be represented as Eq.\ref{eq: ref_flow_loss} where $z^{HR}_{t}=(1-t)z^{HR}+t\epsilon$.
\begin{equation}
\label{eq: ref_flow_loss}
\mathcal{L}_{\psi} = \int_{0}^{1}\mathbb{E}[||v_\psi (z^{HR}_{t},z^{LR},z^{Ref},t)-(\epsilon - z^{HR})||^{2}]dt.
\end{equation}

\begin{figure}[ht]
\centering
\includegraphics[width=0.95\columnwidth]{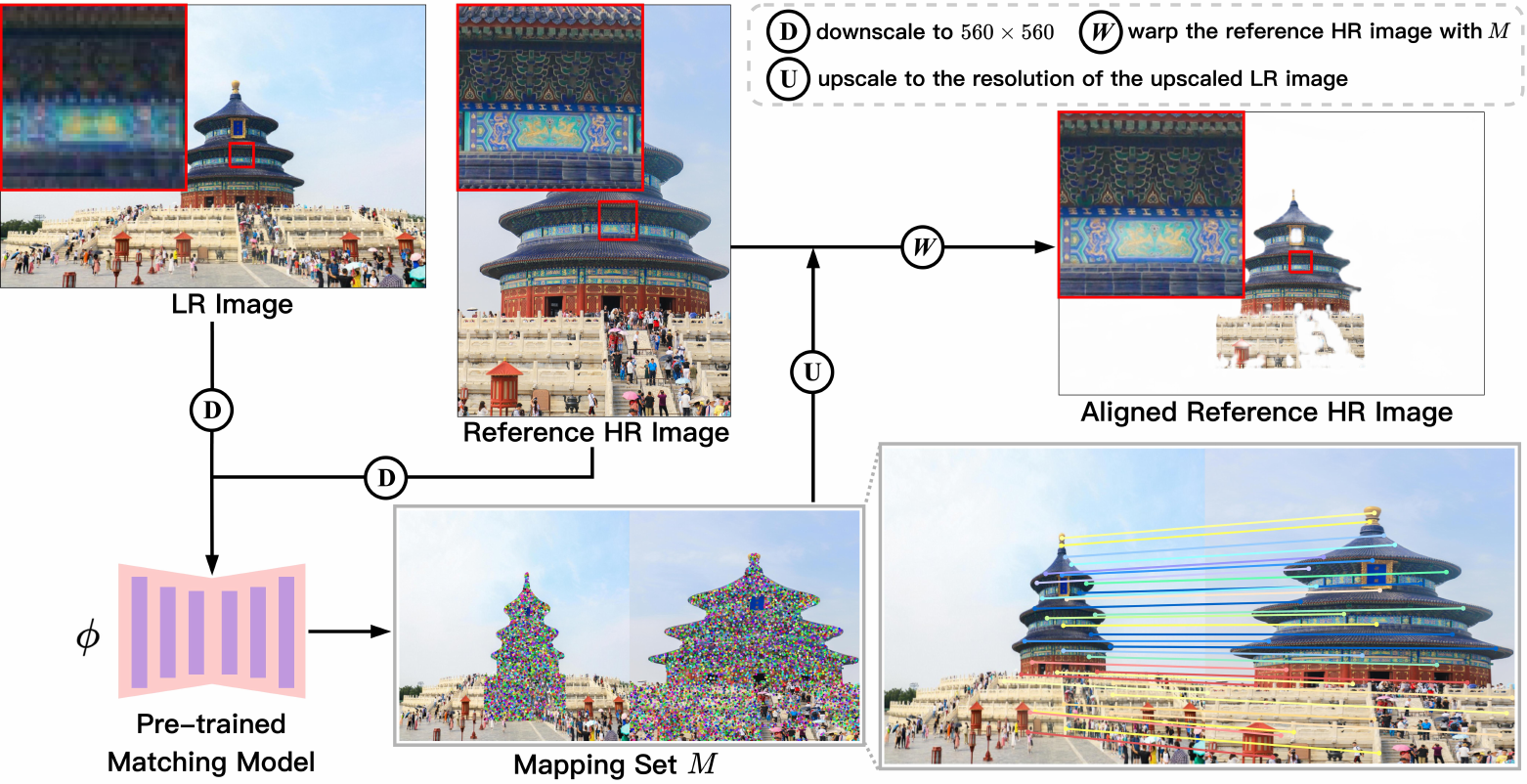} 
\caption{Illustration of the Reference Matching Strategy. We obtain the coarse mapping from the reference HR image to the LR image at a low resolution, and upscale the mapping to the upscaled LR image resolution. Then we warp the reference HR image to align it with the upscaled LR image.}
\label{fig: reference matching}
\end{figure}
\subsection{Reference Matching Strategy}
For Ultra-High-Definition (UHD) images, directly processing the entire image for super-resolution is computationally expensive. To address this, previous SISR methods \cite{wang2024exploiting,  wu2024seesr} typically adopt a tiling strategy, where the input image is divided into smaller tiles, each processed independently and subsequently stitched together to reconstruct the full-resolution image. However, for RefSR, this tiling strategy encounters unique challenges. Due to the inherent differences in scale, perspective, and content between the reference HR image and the LR input image, the tiles from the reference image usually do not correspond directly to the patches of the LR image. To overcome this issue, we propose the \textbf{Reference Matching Strategy} that aligns the reference HR image with the LR image at the pixel level. 

Given the bicubic-upscaled LR image $I^{LR}$ and the reference HR image $I^{Ref}$, we first downscale them to $560\times560$. We then employ a pre-trained matching model $\phi$ to establish a coarse correspondence between the reference HR image and the upscaled LR image, where $(u,v) \in I^{LR}_{down} , (x,y) \in I^{Ref}_{down}$:
\begin{equation}
\label{eq: coarse_matching}
M ,C \leftarrow \mathcal{F}(I^{LR}_{down},I^{Ref}_{down};\phi),
\end{equation}
\begin{equation}
\label{eq: coarse_matching_details}
M(u,v)=(u,v)\mapsto(x,y), C(u,v)\in[0,1]
\end{equation}
Here, $M$ represents the mapping set of points between the  downscaled reference HR image and the  downscaled LR image, and $C$ represents the certainty of each correspondence. Then, we upscale the $M$ and $C$ to the resolution of upscaled LR image. Based on the upscaled $M$, we use $F.grid\_sample$ to warp the corresponding reference HR image to $I^{Ref}_{warped}$. Then we utilize upscaled $C$ to suppress low-confidence textures, as $C \odot I^{Ref}_{warped} + (1-C) \odot mask$, where $\odot$ represents the dot product operation and $mask$ represents a pure white image in Figure \ref{fig: reference matching}. As shown in Figure \ref{fig: reference matching}, by leveraging effective feature matching and alignment techniques, our strategy enables the reference image to provide more accurate and spatially consistent texture information.

\begin{figure}[ht]
\centering
\includegraphics[width=0.95\columnwidth]{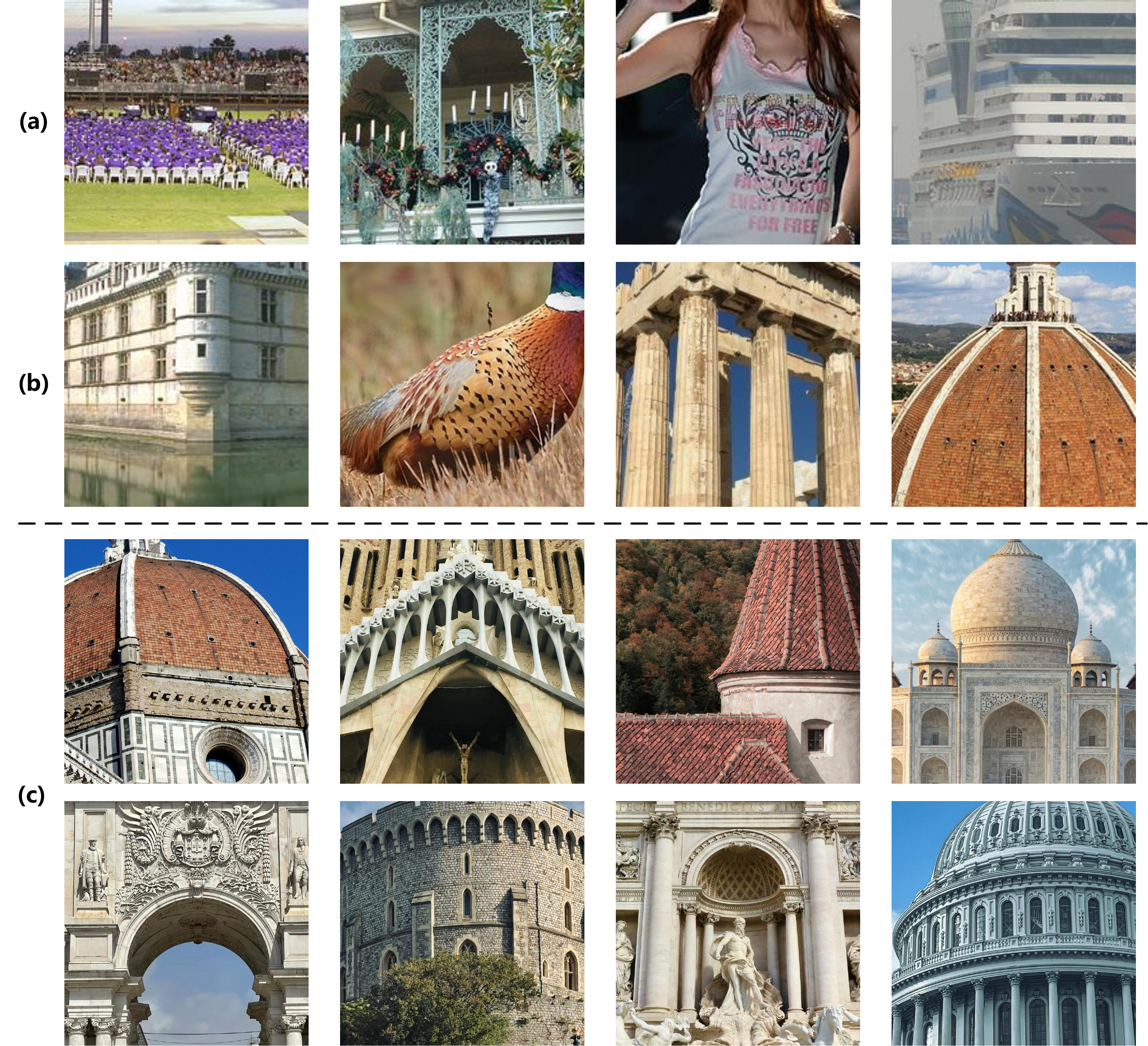} 
\caption{Illustration of the RefSR datasets. (a) represents the CUFED5 dataset \cite{zhang2019image}, (b) represents the WR-SR dataset \cite{jiang2021robust}, and (c) represents our proposed Landmark-4K dataset. We apply a center crop to these images, extracting the central region that accounts for 20\% of the original image area. Notably, the Landmark-4K dataset offers higher resolution and image quality compared to existing RefSR datasets. Please zoom in for a better view.}
\label{fig: datasets}
\end{figure}
\begin{table}[ht]\small
\centering
\setlength{\tabcolsep}{1mm}{
\begin{tabular}{c | c c}
Datasets    & Numbers & Average Resolutions \\
\toprule
CUFED5      & 126     & $418\times418$      \\
WR-SR        & 80      & $770\times770$      \\
Landmark-4K (ours) & 185     & $3295\times3295$   \\
\bottomrule
\end{tabular}}
\caption{A quantitative comparison of the RefSR datasets. The Landmark-4K dataset outperforms the CUFED5 dataset and the WR-SR dataset in terms of both resolution and dataset size.}
\label{tab: datasets}
\end{table}
\subsection{Landmark-4K Dataset}
\begin{figure*}[ht]
\centering
\includegraphics[width=0.95\textwidth]{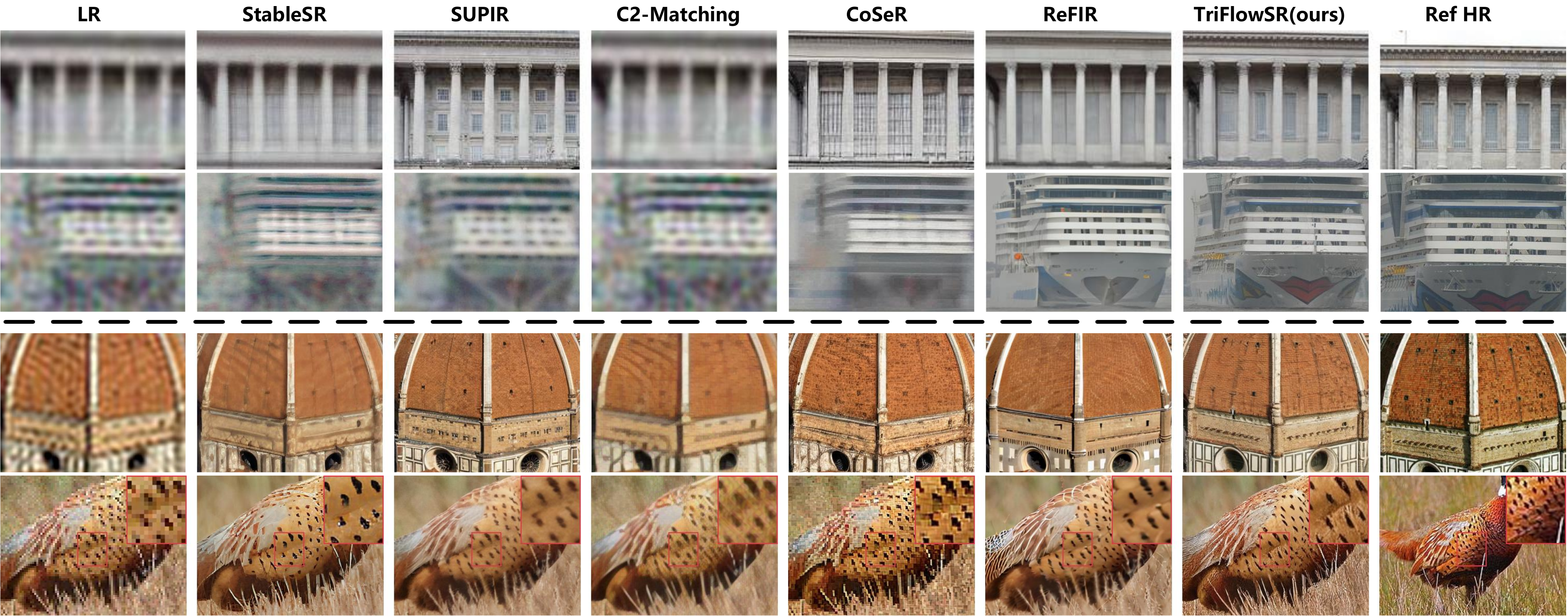} 
\caption{Visual comparisons of the Super-Resolution results by different methods on the CUFED5 dataset and the WR-SR dataset. The top side of the dashed line represents the results on the CUFED 5 dataset, and the bottom side represents the results on the WR-SR dataset.}
\label{fig: CUFED_WRSR_compares}
\end{figure*}
\begin{table*}[ht]\small
\centering
\begin{tabular}{c|ccccc|ccccc}
\toprule
\multirow{2}{*}{Methods}            & \multicolumn{5}{c|}{CUFED5 \cite{zhang2019image}}                                         & \multicolumn{5}{c}{WR-SR \cite{jiang2021robust}}                                                          \\
     & PSNR$\uparrow$	& SSIM$\uparrow$          & LPIPS$\downarrow$         & FID$\downarrow$            & DISTS$\downarrow$         & PSNR$\uparrow$	& SSIM$\uparrow$          & LPIPS$\downarrow$         & FID$\downarrow$           & DISTS$\downarrow$         \\
\midrule
StableSR     &\textbf{20.79} 	& 0.529          & 0.515          & 201.29          & 0.303          &23.17	& 0.597          & 0.409          & 88.85          & 0.246          \\
DiffBIR     &18.43	& 0.361          & 0.508          & 206.33          & 0.246          &21.35	& 0.468          & 0.451          & 102.94         & 0.227          \\
SUPIR       &18.97	& 0.467          & 0.481          & 168.26    & 0.279          &21.82		& 0.517          & 0.381          & \underline{64.67}    & 0.197          \\
TSDSR       &19.18	& 0.491          & \underline{0.327}    & 152.64          & 0.192          &22.04	& 0.577          & \underline{0.333}    & 77.71          & 0.195          \\
\midrule
C2-Matching &\underline{20.77}	& 0.517    & 0.728          & 282.43          & 0.372          &\underline{23.82}	& \underline{0.613}    & 0.658          & 142.79         & 0.318          \\
DATSR       &20.75	& 0.513 & 0.730           & 282.19          & 0.370           &\textbf{23.83}	& \textbf{0.615} & 0.667          & 142.71         & 0.315          \\
CoSeR       &19.94	& 0.503          & 0.393          & 158.70           & 0.220           &22.71	& 0.573          & 0.430           & 112.38         & 0.277          \\
ReFIR       &20.32	& \underline{0.529}    & 0.334          & \underline{134.62}    & \underline{0.186}    &20.99	& 0.532          & 0.373          & 65.48          & \underline{0.195}    \\
\midrule
\textbf{TriFlowSR}   &20.21	& \textbf{0.535} & \textbf{0.275} & \textbf{114.93} & \textbf{0.166} & 22.11	& 0.578          & \textbf{0.313} & \textbf{54.76} & \textbf{0.174} \\
\bottomrule
\end{tabular}
\caption{Quantitative comparisons with other methods on the CUFED5 dataset and the WR-SR dataset. We report PSNR, SSIM, LPIPS, FID and DISTS. The best and second-best results are highlighted in \textbf{bold} and \underline{underlined}. “$\uparrow$" (resp. “$\downarrow$") means the larger (resp. smaller), the better.}
\label{tab: compares public}
\end{table*}
Reference-based Super-Resolution (RefSR) aims to leverage the semantic structures and fine-grained texture details from the reference HR image to guide the reconstruction of low-resolution images, thereby producing more realistic super-resolved results. However, existing RefSR datasets \cite{zhang2019image, jiang2021robust} generally suffer from low resolution and poor image quality, resulting in the reference HR images lacking sufficient detail. This limitation restricts the performance of pre-trained text-to-image (T2I) diffusion models to generate high-fidelity images within the RefSR framework. It is evident that Ultra-High-Definition (UHD) images are inherently more suitable for the RefSR task, as they provide significantly richer structural and textural information for high-quality reconstruction. Therefore, building a high-quality UHD RefSR dataset is of great importance for advancing the development of RefSR methods and unlocking the full capacity of generative models in high-fidelity image restoration.

To construct a high-quality Ultra-High-Definition (UHD) dataset for Reference-based Image Super-Resolution (RefSR), we first collected approximately 1,000 landmark images from publicly available online sources\footnote{pexels.com}. We then performed an initial filtering step by removing all images with a resolution lower than 2K. Subsequently, we resized images with a resolution larger than 4096 by scaling them proportionally to a maximum size of 4096 pixels, in order to prevent excessive image sizes while preserving essential details. We further refined the dataset by removing images with noticeable blur or noise. Lastly, we categorized the images according to landmark types and removed images with significant viewpoint variations within each category to ensure effective semantic and texture correspondence between the reference and target images.
Following this rigorous data curation pipeline, we propose the Landmark-4K dataset, which consists of 185 high-quality landmark images covering 49 landmark categories from the worldwide. Each landmark category contains 3 to 4 images captured from different viewpoints. Specifically, one image from each category is designated for evaluating the self-reference capability of RefSR models, while the remaining images are used for testing their cross-reference generalization ability. The Landmark-4K dataset is designed to serve as a high-quality, challenging, and practical benchmark for advancing research in the field of Reference-based Image Super-resolution.
\begin{figure*}[ht]
\centering
\includegraphics[width=0.95\textwidth]{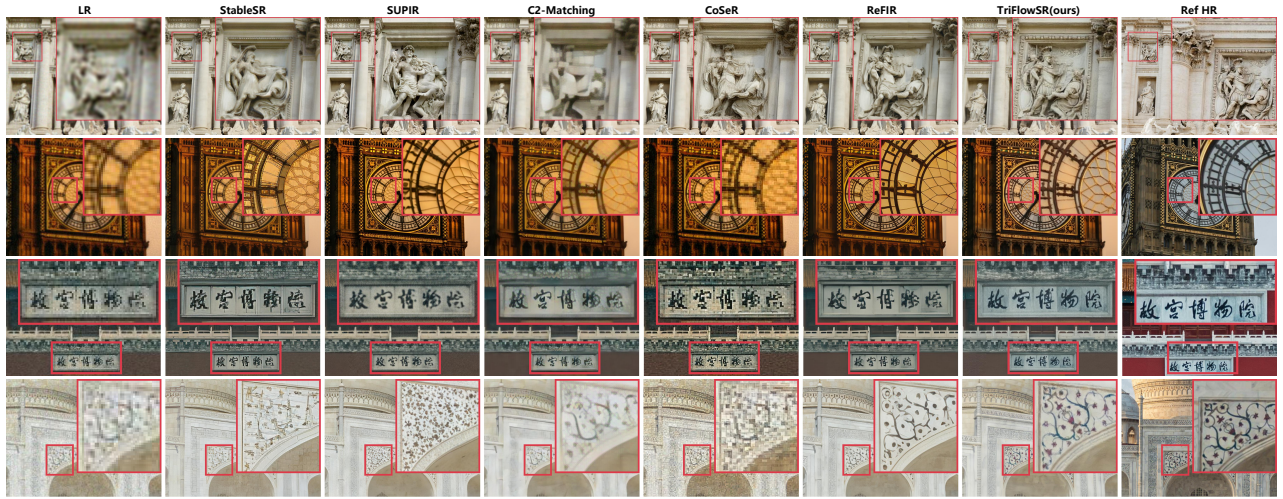} 
\caption{Visual comparisons of the Super-Resolution results by different methods on our proposed Lanmark-4K dataset. Please zoom in for a better view. These images are the results of manual cropping and alignment on the original outputs. In reality, the LR image and the reference HR image may have misalignments in terms of scale, position, and orientation. More visual comparisons are available in the Appendix due to space limitations.}
\label{fig: Landmark4K_compares}
\end{figure*}
\begin{table}[ht]\small
\centering
\begin{tabular}{c|cccc}
\toprule
\multirow{2}{*}{Methods} & \multicolumn{4}{|c}{Landmark-4K}                                   \\
                        & PSNR$\uparrow$          & SSIM$\uparrow$          & LPIPS$\downarrow$         & DISTS$\downarrow$         \\
\midrule
StableSR                &24.90	& 0.722          & 0.308          & 0.153          \\
DiffBIR                 &23.99	& 0.610           & 0.463          & 0.208          \\
SUPIR                   &24.44	& 0.687          & 0.312          & \underline{0.137}    \\
TSDSR                   &23.26	& 0.692          & 0.295          & 0.154          \\
\midrule
C2-Matching             &25.63	& 0.721          & 0.578          & 0.272          \\
DATSR                   &\underline{25.64}	& 0.722          & 0.597          & 0.269          \\
CoSeR                   &24.72	& 0.697          & 0.359          & 0.220           \\
ReFIR                   &25.21	& \underline{0.735}    & \underline{0.280}     & 0.149         \\
\midrule
\textbf{TriFlowSR}               &\textbf{25.89}	& \textbf{0.776} & \textbf{0.230} & \textbf{0.115} \\
\bottomrule
\end{tabular}
\caption{Quantitative comparisons with other methods on our proposed Landmark-4K dataset. We report PSNR, SSIM, LPIPS and DISTS. The best and second-best results are highlighted in \textbf{bold} and \underline{underlined}. “$\uparrow$" (resp. “$\downarrow$") means the larger (resp. smaller), the better.}
\label{tab: compares landmark}
\end{table}
\section{Evaluation}
\subsection{Datasets and Baselines.}
To evaluate our method, we adopt three
benchmarks: CUFED5 \cite{zhang2019image} testing dataset, WR-SR \cite{jiang2021robust} dataset and our proposed Landmark-4K dataset. CUFED5 testing dataset consists of 126 image pairs, and each input image has 5 reference images with different similarity levels. Following previous methods \cite{yang2020learning, lu2021masa, xia2022coarse, cao2022reference}, we use the most similar image as the reference HR image. WR-SR dataset consists of 80 images, and each image has one reference image searched through Google Image. The Landmark-4K dataset consists of 185 high-quality landmark images covering 49 landmark categories from the worldwide, with each image having a corresponding high-quality reference HR image from the same landmark scenario. Following previous methods \cite{guo2024refir, zhang2024reference}, we use the Real-ESRGAN \cite{wang2021real} degradation pipeline (same configuration as StableSR \cite{wang2024exploiting}) with $\times4$ down-sampling scale to generate the real-world degraded images.
To validate the performance of our model, we compare the proposed method with both the popular SISR and RefSR methods. The SISR methods include StableSR \cite{wang2024exploiting}, DiffBIR \cite{lin2024diffbir}, SUPIR \cite{yu2024scaling} and TSDSR \cite{dong2025tsd}. The RefSR methods include C2-Matching \cite{jiang2021robust}, DATSR \cite{cao2022reference}, CoSeR \cite{sun2024coser} and ReFIR \cite{guo2024refir}. 

\subsection{Implementation Details and Evaluation Metrics.}
In the experiment, we train the LR branch based on the LSDIR dataset \cite{li2023lsdir} and the first 10K face images from the FFHQ \cite{karras2019style} dataset in the first stage, following common Super-Resolution settings \cite{wang2024exploiting, dong2025tsd}. In the second stage, we train the Reference HR branch based on the DL3DV \cite{ling2024dl3dv} dataset and the Inter4K \cite{stergiou2022adapool} dataset. We use Stable Diffusion 3 \cite{esser2024scaling} as the pre-trained diffusion backbone for the SR Branch, and kept it frozen throughout all training stages. We use Roma \cite{edstedt2024roma} as the pre-trained matching model $\phi$ for the Reference Matching Strategy. More training details are provided in the Appendix.
To verify the performance of different models, we conduct quantitative evaluations, including both structural metrics, perceptual metrics, and distribution consistency metrics. We use the Peak Signal-to-Noise Ratio (PSNR), Structural Similarity Index Measure (SSIM) \cite{wang2004image}, Learned Perceptual Image Patch Similarity (LPIPS) \cite{zhang2018unreasonable} and  Deep Image Structure and Texture Similarity (DISTS) \cite{ding2020image} to assess the quality of the generated images. We employ the Frechet Inception Distance (FID) \cite{heusel2017gans} to evaluate the realism of the results. Since FID involves downsampling before calculation and is not suitable for UHD images, we did not quantify the FID metric on the Landmark-4K dataset.

\subsection{Qualitative and Quantitative Results.}
As shown in Table \ref{tab: compares public}, our method achieves the best SSIM, LPIPS, FID, and DISTS on the public CUFED dataset. On the public WR-SR dataset, our method obtains the best LPIPS, FID and DISTS. Since the reference HR images in WR-SR dataset are retrieved from Google Image Search, their alignment with the LR images is relatively poor (as illustrated in Figure \ref{fig: CUFED_WRSR_compares}). As a result, our method may yield slightly lower PSNR and SSIM compared to some existing methods. However, our method significantly outperforms them in perceptual quality and distribution consistency metrics, including LPIPS, FID and DISTS. Furthermore, as shown in Table \ref{tab: compares landmark}, our method achieves the best PSNR, SSIM, LPIPS, and DISTS on our proposed Landmark-4K dataset. As shown in Figure \ref{fig: CUFED_WRSR_compares} and Figure \ref{fig: Landmark4K_compares}, benefiting from explicitly pattern matching the LR image and the reference HR image, our approach better leverages the semantic and texture information from the reference HR image compared to other methods, resulting in more realistic textures and finer details.

\section{Ablation Study}
\subsection{Effectiveness of the Components}
Notably, the LR and Ref features are only concatenated during cross-attention operations, making them completely decoupled. First, we attempt to perform inference with only the LR branch and SR branch, which degenerates our model into a SISR model. Then, we introduce the Reference branch, but only utilize the relative position of the LR image tiles to map to the reference HR image tiles. Next, we use the pre-trained matching model to obtain the matching map and simply resize the corresponding reference region to the input tile size without using warping to align the images. Finally, we apply the warping operation to align the LR image with the reference HR image. As shown in Table \ref{tab: ablation}, compared to pure SISR, introducing the Reference branch can significantly improve PSNR, SSIM, LPIPS, and DISTS. Introducing the Reference Matching Strategy can solve the problem of incorrect tile retrieval due to scale mismatch and positional differences between the LR image and the reference HR image. Visual comparisons are provided in the Appendix due to space constraints.

\begin{table}[ht]\small
\centering
\begin{tabular}{ccc|cccc}
\toprule
RB & M & W & PSNR↑ & SSIM↑ & LPIPS↓ & DISTS↓\\
\midrule
           &              &         & 24.07 & 0.664 & 0.355  & 0.166  \\
\checkmark          &              &         &\underline{25.67}       &\underline{0.756}      &0.241        &\underline{0.113}        \\
\checkmark          & \checkmark            &         & 25.37 & 0.750  & \underline{0.239}  & \textbf{0.110}   \\
\checkmark          & \checkmark            & \checkmark       & \textbf{25.89} & \textbf{0.776} & \textbf{0.230}  & 0.115  \\
\bottomrule
\end{tabular}
\caption{Ablation study on our proposed Landmark-4K dataset. We report PSNR, SSIM, LPIPS and DISTS. RB represents the Reference branch, M represents the Matching operation and W represents the Warping operation. The best and second-best results are highlighted in \textbf{bold} and \underline{underlined}. “$\uparrow$" (resp. “$\downarrow$") means the larger (resp. smaller), the better.}
\label{tab: ablation}
\end{table}
\begin{table}[ht]\small
\centering
\begin{tabular}{c|cccc}
\toprule
$kscale$ & PSNR↑ & SSIM↑ & LPIPS↓ & DISTS↓\\
\midrule
0        & 24.07 & 0.664 & 0.355  & 0.166  \\
0.2      & 24.36 & 0.712 & 0.299  & 0.181  \\
0.4      & 24.57 & 0.709 & 0.287  & 0.159  \\
0.6      & 24.65 & 0.710 & 0.285  & 0.150  \\
0.8      & \underline{24.79} & \underline{0.716} & \underline{0.280}  & \underline{0.145}  \\
1        & \textbf{25.89} & \textbf{0.776} & \textbf{0.230}  & \textbf{0.115}  \\
\bottomrule
\end{tabular}
\caption{Quantitative comparisons about the control scale ($kscale$) of the Reference branch on our proposed Landmark-4K dataset. We report PSNR, SSIM, LPIPS and DISTS. The best and second-best results are highlighted in \textbf{bold} and \underline{underlined}. “$\uparrow$" (resp. “$\downarrow$") means the larger (resp. smaller), the better.}
\label{tab: ref_control}
\end{table}
\subsection{Control Scale of the Reference branch}
Similar to ControlNet \cite{zhang2023adding} and IP-adapter \cite{ye2023ip}, TriFlow can also control the influence of the reference branch by adjusting the weight of the attention map. Specifically, during the Pat-Ref Attention operation, we can control the attention weight corresponding to the reference branch feature by multiplying $K^{Ref}$ by a coefficient $kscale$, i.e., $K^{Ref} \mapsto kscale \times K^{Ref}$. Since we only performed concatenation on the $K$ and $V$ features, directly controlling the weight of $K^{ref}$ can control the corresponding attention weight $Q [kscale \times K^{ref}]^{T}$ of $V^{ref}$, thereby achieving the result of controlling the influence of the reference branch. As shown in Table \ref{tab: ref_control}, when $kscale=1$, ours is a RefSR model. When the value of $kscale$ is between $0 - 1$, we can consider it as a trade-off between SISR and RefSR. As $kscale$ gradually decreases to zero, ours gradually degenerates into a SISR model. More perspective comparison results are detailed in the Appendix due to space constraints. 

\section{Conclusion}
In this paper, we address the problem of Ultra-High-Definition Reference-Based Landmark Image Super-Resolution. To this end, we propose TriflowSR, a novel framework that explicitly performs pattern matching between LR and reference HR features. To address the issues of low quality and relatively low resolution in existing RefSR datasets, we propose the first UHD RefSR dataset based on landmark scenes and design a Reference Matching Strategy to match corresponding tiles. We successfully implement a diffusion-based RefSR pipeline in UHD landmark scenarios under the real-world degradation. Experimental results show that our approach can better utilize the semantic and texture information of the reference HR image than previous methods, resulting in more realistic texture and details.

\bibliography{aaai2026}

\begin{thebibliography}{52}
\providecommand{\natexlab}[1]{#1}

\bibitem[{Cao et~al.(2022)Cao, Liang, Zhang, Li, Zhang, Wang, and Gool}]{cao2022reference}
Cao, J.; Liang, J.; Zhang, K.; Li, Y.; Zhang, Y.; Wang, W.; and Gool, L.~V. 2022.
\newblock Reference-based image super-resolution with deformable attention transformer.
\newblock In \emph{European conference on computer vision}, 325--342. Springer.

\bibitem[{Dai et~al.(2017)Dai, Qi, Xiong, Li, Zhang, Hu, and Wei}]{dai2017deformable}
Dai, J.; Qi, H.; Xiong, Y.; Li, Y.; Zhang, G.; Hu, H.; and Wei, Y. 2017.
\newblock Deformable convolutional networks.
\newblock In \emph{Proceedings of the IEEE international conference on computer vision}, 764--773.

\bibitem[{Ding et~al.(2020)Ding, Ma, Wang, and Simoncelli}]{ding2020image}
Ding, K.; Ma, K.; Wang, S.; and Simoncelli, E.~P. 2020.
\newblock Image quality assessment: Unifying structure and texture similarity.
\newblock \emph{IEEE transactions on pattern analysis and machine intelligence}, 44(5): 2567--2581.

\bibitem[{Dong et~al.(2014)Dong, Loy, He, and Tang}]{dong2014learning}
Dong, C.; Loy, C.~C.; He, K.; and Tang, X. 2014.
\newblock Learning a deep convolutional network for image super-resolution.
\newblock In \emph{Computer Vision--ECCV 2014: 13th European Conference, Zurich, Switzerland, September 6-12, 2014, Proceedings, Part IV 13}, 184--199. Springer.

\bibitem[{Dong et~al.(2025)Dong, Fan, Guo, Wang, Zhang, Chen, Luo, and Zou}]{dong2025tsd}
Dong, L.; Fan, Q.; Guo, Y.; Wang, Z.; Zhang, Q.; Chen, J.; Luo, Y.; and Zou, C. 2025.
\newblock Tsd-sr: One-step diffusion with target score distillation for real-world image super-resolution.
\newblock In \emph{Proceedings of the Computer Vision and Pattern Recognition Conference}, 23174--23184.

\bibitem[{Edstedt et~al.(2024)Edstedt, Sun, B{\"o}kman, Wadenb{\"a}ck, and Felsberg}]{edstedt2024roma}
Edstedt, J.; Sun, Q.; B{\"o}kman, G.; Wadenb{\"a}ck, M.; and Felsberg, M. 2024.
\newblock Roma: Robust dense feature matching.
\newblock In \emph{Proceedings of the IEEE/CVF Conference on Computer Vision and Pattern Recognition}, 19790--19800.

\bibitem[{Esser et~al.(2024)Esser, Kulal, Blattmann, Entezari, M{\"u}ller, Saini, Levi, Lorenz, Sauer, Boesel et~al.}]{esser2024scaling}
Esser, P.; Kulal, S.; Blattmann, A.; Entezari, R.; M{\"u}ller, J.; Saini, H.; Levi, Y.; Lorenz, D.; Sauer, A.; Boesel, F.; et~al. 2024.
\newblock Scaling rectified flow transformers for high-resolution image synthesis.
\newblock In \emph{Forty-first international conference on machine learning}.

\bibitem[{Esser, Rombach, and Ommer(2021)}]{esser2021taming}
Esser, P.; Rombach, R.; and Ommer, B. 2021.
\newblock Taming transformers for high-resolution image synthesis.
\newblock In \emph{Proceedings of the IEEE/CVF conference on computer vision and pattern recognition}, 12873--12883.

\bibitem[{Goodfellow et~al.(2020)Goodfellow, Pouget-Abadie, Mirza, Xu, Warde-Farley, Ozair, Courville, and Bengio}]{goodfellow2020generative}
Goodfellow, I.; Pouget-Abadie, J.; Mirza, M.; Xu, B.; Warde-Farley, D.; Ozair, S.; Courville, A.; and Bengio, Y. 2020.
\newblock Generative adversarial networks.
\newblock \emph{Communications of the ACM}, 63(11): 139--144.

\bibitem[{Guo et~al.(2024)Guo, Dai, Ouyang, Zhang, Zha, Chen, and Xia}]{guo2024refir}
Guo, H.; Dai, T.; Ouyang, Z.; Zhang, T.; Zha, Y.; Chen, B.; and Xia, S.-t. 2024.
\newblock Refir: Grounding large restoration models with retrieval augmentation.
\newblock \emph{Advances in Neural Information Processing Systems}, 37: 46593--46621.

\bibitem[{Heusel et~al.(2017)Heusel, Ramsauer, Unterthiner, Nessler, and Hochreiter}]{heusel2017gans}
Heusel, M.; Ramsauer, H.; Unterthiner, T.; Nessler, B.; and Hochreiter, S. 2017.
\newblock Gans trained by a two time-scale update rule converge to a local nash equilibrium.
\newblock \emph{Advances in neural information processing systems}, 30.

\bibitem[{Ho, Jain, and Abbeel(2020)}]{ho2020denoising}
Ho, J.; Jain, A.; and Abbeel, P. 2020.
\newblock Denoising diffusion probabilistic models.
\newblock \emph{Advances in neural information processing systems}, 33: 6840--6851.

\bibitem[{Hu(2024)}]{hu2024animate}
Hu, L. 2024.
\newblock Animate anyone: Consistent and controllable image-to-video synthesis for character animation.
\newblock In \emph{Proceedings of the IEEE/CVF Conference on Computer Vision and Pattern Recognition}, 8153--8163.

\bibitem[{Huang et~al.(2022)Huang, Zhang, Fu, Chen, Zhang, Wang, and He}]{huang2022task}
Huang, Y.; Zhang, X.; Fu, Y.; Chen, S.; Zhang, Y.; Wang, Y.-F.; and He, D. 2022.
\newblock Task decoupled framework for reference-based super-resolution.
\newblock In \emph{Proceedings of the IEEE/CVF Conference on Computer Vision and Pattern Recognition}, 5931--5940.

\bibitem[{Jiang et~al.(2021)Jiang, Chan, Wang, Loy, and Liu}]{jiang2021robust}
Jiang, Y.; Chan, K.~C.; Wang, X.; Loy, C.~C.; and Liu, Z. 2021.
\newblock Robust reference-based super-resolution via c2-matching.
\newblock In \emph{Proceedings of the IEEE/CVF Conference on Computer Vision and Pattern Recognition}, 2103--2112.

\bibitem[{Karras, Laine, and Aila(2019)}]{karras2019style}
Karras, T.; Laine, S.; and Aila, T. 2019.
\newblock A style-based generator architecture for generative adversarial networks.
\newblock In \emph{Proceedings of the IEEE/CVF conference on computer vision and pattern recognition}, 4401--4410.

\bibitem[{Kim, Lee, and Lee(2016{\natexlab{a}})}]{kim2016accurate}
Kim, J.; Lee, J.~K.; and Lee, K.~M. 2016{\natexlab{a}}.
\newblock Accurate image super-resolution using very deep convolutional networks.
\newblock In \emph{Proceedings of the IEEE conference on computer vision and pattern recognition}, 1646--1654.

\bibitem[{Kim, Lee, and Lee(2016{\natexlab{b}})}]{kim2016deeply}
Kim, J.; Lee, J.~K.; and Lee, K.~M. 2016{\natexlab{b}}.
\newblock Deeply-recursive convolutional network for image super-resolution.
\newblock In \emph{Proceedings of the IEEE conference on computer vision and pattern recognition}, 1637--1645.

\bibitem[{Lai et~al.(2017)Lai, Huang, Ahuja, and Yang}]{lai2017deep}
Lai, W.-S.; Huang, J.-B.; Ahuja, N.; and Yang, M.-H. 2017.
\newblock Deep laplacian pyramid networks for fast and accurate super-resolution.
\newblock In \emph{Proceedings of the IEEE conference on computer vision and pattern recognition}, 624--632.

\bibitem[{Ledig et~al.(2017)Ledig, Theis, Husz{\'a}r, Caballero, Cunningham, Acosta, Aitken, Tejani, Totz, Wang et~al.}]{ledig2017photo}
Ledig, C.; Theis, L.; Husz{\'a}r, F.; Caballero, J.; Cunningham, A.; Acosta, A.; Aitken, A.; Tejani, A.; Totz, J.; Wang, Z.; et~al. 2017.
\newblock Photo-realistic single image super-resolution using a generative adversarial network.
\newblock In \emph{Proceedings of the IEEE conference on computer vision and pattern recognition}, 4681--4690.

\bibitem[{Li et~al.(2023)Li, Zhang, Liang, Cao, Liu, Gong, Zhang, Tang, Liu, Demandolx et~al.}]{li2023lsdir}
Li, Y.; Zhang, K.; Liang, J.; Cao, J.; Liu, C.; Gong, R.; Zhang, Y.; Tang, H.; Liu, Y.; Demandolx, D.; et~al. 2023.
\newblock Lsdir: A large scale dataset for image restoration.
\newblock In \emph{Proceedings of the IEEE/CVF Conference on Computer Vision and Pattern Recognition}, 1775--1787.

\bibitem[{Lim et~al.(2017)Lim, Son, Kim, Nah, and Mu~Lee}]{lim2017enhanced}
Lim, B.; Son, S.; Kim, H.; Nah, S.; and Mu~Lee, K. 2017.
\newblock Enhanced deep residual networks for single image super-resolution.
\newblock In \emph{Proceedings of the IEEE conference on computer vision and pattern recognition workshops}, 136--144.

\bibitem[{Lin et~al.(2024)Lin, He, Chen, Lyu, Dai, Yu, Qiao, Ouyang, and Dong}]{lin2024diffbir}
Lin, X.; He, J.; Chen, Z.; Lyu, Z.; Dai, B.; Yu, F.; Qiao, Y.; Ouyang, W.; and Dong, C. 2024.
\newblock Diffbir: Toward blind image restoration with generative diffusion prior.
\newblock In \emph{European Conference on Computer Vision}, 430--448. Springer.

\bibitem[{Ling et~al.(2024)Ling, Sheng, Tu, Zhao, Xin, Wan, Yu, Guo, Yu, Lu et~al.}]{ling2024dl3dv}
Ling, L.; Sheng, Y.; Tu, Z.; Zhao, W.; Xin, C.; Wan, K.; Yu, L.; Guo, Q.; Yu, Z.; Lu, Y.; et~al. 2024.
\newblock Dl3dv-10k: A large-scale scene dataset for deep learning-based 3d vision.
\newblock In \emph{Proceedings of the IEEE/CVF Conference on Computer Vision and Pattern Recognition}, 22160--22169.

\bibitem[{Liu, Gong, and Liu(2022)}]{liu2022flow}
Liu, X.; Gong, C.; and Liu, Q. 2022.
\newblock Flow straight and fast: Learning to generate and transfer data with rectified flow.
\newblock \emph{arXiv preprint arXiv:2209.03003}.

\bibitem[{Liu et~al.(2021)Liu, Lin, Cao, Hu, Wei, Zhang, Lin, and Guo}]{liu2021swin}
Liu, Z.; Lin, Y.; Cao, Y.; Hu, H.; Wei, Y.; Zhang, Z.; Lin, S.; and Guo, B. 2021.
\newblock Swin transformer: Hierarchical vision transformer using shifted windows.
\newblock In \emph{Proceedings of the IEEE/CVF international conference on computer vision}, 10012--10022.

\bibitem[{Lu et~al.(2021)Lu, Li, Tao, Lu, and Jia}]{lu2021masa}
Lu, L.; Li, W.; Tao, X.; Lu, J.; and Jia, J. 2021.
\newblock Masa-sr: Matching acceleration and spatial adaptation for reference-based image super-resolution.
\newblock In \emph{Proceedings of the IEEE/CVF Conference on Computer Vision and Pattern Recognition}, 6368--6377.

\bibitem[{Mou et~al.(2024)Mou, Wang, Xie, Wu, Zhang, Qi, and Shan}]{mou2024t2i}
Mou, C.; Wang, X.; Xie, L.; Wu, Y.; Zhang, J.; Qi, Z.; and Shan, Y. 2024.
\newblock T2i-adapter: Learning adapters to dig out more controllable ability for text-to-image diffusion models.
\newblock In \emph{Proceedings of the AAAI conference on artificial intelligence}, volume~38, 4296--4304.

\bibitem[{Rombach et~al.(2022)Rombach, Blattmann, Lorenz, Esser, and Ommer}]{rombach2022high}
Rombach, R.; Blattmann, A.; Lorenz, D.; Esser, P.; and Ommer, B. 2022.
\newblock High-resolution image synthesis with latent diffusion models.
\newblock In \emph{Proceedings of the IEEE/CVF conference on computer vision and pattern recognition}, 10684--10695.

\bibitem[{Shim, Park, and Kweon(2020)}]{shim2020robust}
Shim, G.; Park, J.; and Kweon, I.~S. 2020.
\newblock Robust reference-based super-resolution with similarity-aware deformable convolution.
\newblock In \emph{Proceedings of the IEEE/CVF conference on computer vision and pattern recognition}, 8425--8434.

\bibitem[{Song, Meng, and Ermon(2020)}]{song2020denoising}
Song, J.; Meng, C.; and Ermon, S. 2020.
\newblock Denoising diffusion implicit models.
\newblock \emph{arXiv preprint arXiv:2010.02502}.

\bibitem[{Song et~al.(2020)Song, Sohl-Dickstein, Kingma, Kumar, Ermon, and Poole}]{song2020score}
Song, Y.; Sohl-Dickstein, J.; Kingma, D.~P.; Kumar, A.; Ermon, S.; and Poole, B. 2020.
\newblock Score-based generative modeling through stochastic differential equations.
\newblock \emph{arXiv preprint arXiv:2011.13456}.

\bibitem[{Stergiou and Poppe(2022)}]{stergiou2022adapool}
Stergiou, A.; and Poppe, R. 2022.
\newblock Adapool: Exponential adaptive pooling for information-retaining downsampling.
\newblock \emph{IEEE Transactions on Image Processing}, 32: 251--266.

\bibitem[{Sun et~al.(2024)Sun, Li, Liu, Chen, Pei, Zou, Yan, and Yang}]{sun2024coser}
Sun, H.; Li, W.; Liu, J.; Chen, H.; Pei, R.; Zou, X.; Yan, Y.; and Yang, Y. 2024.
\newblock Coser: Bridging image and language for cognitive super-resolution.
\newblock In \emph{Proceedings of the IEEE/CVF Conference on Computer Vision and Pattern Recognition}, 25868--25878.

\bibitem[{Van Den~Oord, Vinyals et~al.(2017)}]{van2017neural}
Van Den~Oord, A.; Vinyals, O.; et~al. 2017.
\newblock Neural discrete representation learning.
\newblock \emph{Advances in neural information processing systems}, 30.

\bibitem[{Wang et~al.(2024)Wang, Yue, Zhou, Chan, and Loy}]{wang2024exploiting}
Wang, J.; Yue, Z.; Zhou, S.; Chan, K.~C.; and Loy, C.~C. 2024.
\newblock Exploiting diffusion prior for real-world image super-resolution.
\newblock \emph{International Journal of Computer Vision}, 132(12): 5929--5949.

\bibitem[{Wang et~al.(2021)Wang, Xie, Dong, and Shan}]{wang2021real}
Wang, X.; Xie, L.; Dong, C.; and Shan, Y. 2021.
\newblock Real-esrgan: Training real-world blind super-resolution with pure synthetic data.
\newblock In \emph{Proceedings of the IEEE/CVF international conference on computer vision}, 1905--1914.

\bibitem[{Wang et~al.(2004)Wang, Bovik, Sheikh, and Simoncelli}]{wang2004image}
Wang, Z.; Bovik, A.~C.; Sheikh, H.~R.; and Simoncelli, E.~P. 2004.
\newblock Image quality assessment: from error visibility to structural similarity.
\newblock \emph{IEEE transactions on image processing}, 13(4): 600--612.

\bibitem[{Wu et~al.(2024{\natexlab{a}})Wu, Sun, Ma, and Zhang}]{wu2024one}
Wu, R.; Sun, L.; Ma, Z.; and Zhang, L. 2024{\natexlab{a}}.
\newblock One-step effective diffusion network for real-world image super-resolution.
\newblock \emph{Advances in Neural Information Processing Systems}, 37: 92529--92553.

\bibitem[{Wu et~al.(2024{\natexlab{b}})Wu, Yang, Sun, Zhang, Li, and Zhang}]{wu2024seesr}
Wu, R.; Yang, T.; Sun, L.; Zhang, Z.; Li, S.; and Zhang, L. 2024{\natexlab{b}}.
\newblock Seesr: Towards semantics-aware real-world image super-resolution.
\newblock In \emph{Proceedings of the IEEE/CVF conference on computer vision and pattern recognition}, 25456--25467.

\bibitem[{Xia et~al.(2022)Xia, Tian, Hang, Yang, Liao, and Zhou}]{xia2022coarse}
Xia, B.; Tian, Y.; Hang, Y.; Yang, W.; Liao, Q.; and Zhou, J. 2022.
\newblock Coarse-to-fine embedded patchmatch and multi-scale dynamic aggregation for reference-based super-resolution.
\newblock In \emph{Proceedings of the AAAI Conference on Artificial Intelligence}, volume~36, 2768--2776.

\bibitem[{Yang et~al.(2020)Yang, Yang, Fu, Lu, and Guo}]{yang2020learning}
Yang, F.; Yang, H.; Fu, J.; Lu, H.; and Guo, B. 2020.
\newblock Learning texture transformer network for image super-resolution.
\newblock In \emph{Proceedings of the IEEE/CVF conference on computer vision and pattern recognition}, 5791--5800.

\bibitem[{Ye et~al.(2023)Ye, Zhang, Liu, Han, and Yang}]{ye2023ip}
Ye, H.; Zhang, J.; Liu, S.; Han, X.; and Yang, W. 2023.
\newblock Ip-adapter: Text compatible image prompt adapter for text-to-image diffusion models.
\newblock \emph{arXiv preprint arXiv:2308.06721}.

\bibitem[{Yu et~al.(2024)Yu, Gu, Li, Hu, Kong, Wang, He, Qiao, and Dong}]{yu2024scaling}
Yu, F.; Gu, J.; Li, Z.; Hu, J.; Kong, X.; Wang, X.; He, J.; Qiao, Y.; and Dong, C. 2024.
\newblock Scaling up to excellence: Practicing model scaling for photo-realistic image restoration in the wild.
\newblock In \emph{Proceedings of the IEEE/CVF Conference on Computer Vision and Pattern Recognition}, 25669--25680.

\bibitem[{Zhang et~al.(2021)Zhang, Liang, Van~Gool, and Timofte}]{zhang2021designing}
Zhang, K.; Liang, J.; Van~Gool, L.; and Timofte, R. 2021.
\newblock Designing a practical degradation model for deep blind image super-resolution.
\newblock In \emph{Proceedings of the IEEE/CVF international conference on computer vision}, 4791--4800.

\bibitem[{Zhang et~al.(2023)Zhang, Li, He, Li, Ding, and Zhang}]{zhang2023lmr}
Zhang, L.; Li, X.; He, D.; Li, F.; Ding, E.; and Zhang, Z. 2023.
\newblock LMR: a large-scale multi-reference dataset for reference-based super-resolution.
\newblock In \emph{Proceedings of the IEEE/CVF International Conference on Computer Vision}, 13118--13127.

\bibitem[{Zhang, Rao, and Agrawala(2023)}]{zhang2023adding}
Zhang, L.; Rao, A.; and Agrawala, M. 2023.
\newblock Adding conditional control to text-to-image diffusion models.
\newblock In \emph{Proceedings of the IEEE/CVF international conference on computer vision}, 3836--3847.

\bibitem[{Zhang et~al.(2018{\natexlab{a}})Zhang, Isola, Efros, Shechtman, and Wang}]{zhang2018unreasonable}
Zhang, R.; Isola, P.; Efros, A.~A.; Shechtman, E.; and Wang, O. 2018{\natexlab{a}}.
\newblock The unreasonable effectiveness of deep features as a perceptual metric.
\newblock In \emph{Proceedings of the IEEE conference on computer vision and pattern recognition}, 586--595.

\bibitem[{Zhang et~al.(2018{\natexlab{b}})Zhang, Li, Li, Wang, Zhong, and Fu}]{zhang2018image}
Zhang, Y.; Li, K.; Li, K.; Wang, L.; Zhong, B.; and Fu, Y. 2018{\natexlab{b}}.
\newblock Image super-resolution using very deep residual channel attention networks.
\newblock In \emph{Proceedings of the European conference on computer vision (ECCV)}, 286--301.

\bibitem[{Zhang et~al.(2024)Zhang, Yang, Chandler, and Mou}]{zhang2024reference}
Zhang, Y.; Yang, Q.; Chandler, D.~M.; and Mou, X. 2024.
\newblock Reference-Based Multi-Stage Progressive Restoration for Multi-Degraded Images.
\newblock \emph{IEEE Transactions on Image Processing}.

\bibitem[{Zhang et~al.(2019)Zhang, Wang, Lin, and Qi}]{zhang2019image}
Zhang, Z.; Wang, Z.; Lin, Z.; and Qi, H. 2019.
\newblock Image super-resolution by neural texture transfer.
\newblock In \emph{Proceedings of the IEEE/CVF conference on computer vision and pattern recognition}, 7982--7991.

\bibitem[{Zheng et~al.(2018)Zheng, Ji, Wang, Liu, and Fang}]{zheng2018crossnet}
Zheng, H.; Ji, M.; Wang, H.; Liu, Y.; and Fang, L. 2018.
\newblock Crossnet: An end-to-end reference-based super resolution network using cross-scale warping.
\newblock In \emph{Proceedings of the European conference on computer vision (ECCV)}, 88--104.

\end{thebibliography}

\newpage
\appendix
\section{Appendix}

\subsection{Detail Experimental Setting}
In the experiment, we train the LR branch based on the LSDIR dataset \cite{li2023lsdir} and the first 10K face images from the FFHQ \cite{karras2019style} dataset in the first stage, following common Super-Resolution settings \cite{wang2024exploiting, dong2025tsd}. The $I_{LR}$ is upsampled to the desired size using bicubic interpolation during the pre-processing stage, following common practice. In the second stage, we train the Reference HR branch based on the DL3DV \cite{ling2024dl3dv} dataset and the Inter4K \cite{stergiou2022adapool} dataset. Both stages of training are conducted on 4 NVIDIA H20 GPUs. We utilized the AdamW optimizer with a learning rate set to $5e^{-5}$. We use Stable Diffusion 3 \cite{esser2024scaling} as the pre-trained diffusion backbone for the SR Branch, and kept it frozen throughout all training stages. We use Roma \cite{edstedt2024roma} as the pre-trained matching model $\phi$ for the Reference Matching Strategy. In practical implementation, since the mask can be any available image, we will use the result of the SISR (Single Image Super-Resolution) model as the actual mask used.
During the two-stage training phase, we employed mixed training with resolutions of $1024\times1024$ and $512\times512$ to enhance the model's ability to handle images of varying resolutions. Following previous methods \cite{yang2020learning, lu2021masa, xia2022coarse, cao2022reference}, we used the second-order degradation model from Real-ESRGAN \cite{wang2021real} (with the same configuration as StableSR \cite{wang2024exploiting}) with a $\times4$ down-sampling scale to generate real-world degraded images.
Simultaneously, we applied data augmentation techniques such as random flipping, random cropping, ColorJitter, and Homography transformation to improve the model's robustness.
During the inference phase, for the CUFED5 dataset, we padded both the LR image and the reference HR image to a size of $512\times512$ to achieve alignment. For the WR-SR dataset, we padded the images to a square shape with the maximum width and height to align them. For the Landmark-4K dataset, we employed a sliding window approach with a tile size of $1024$ and a tile step of $256$ for tile-based inference, and used the c2-blending strategy to stitch the results back to the original size.
We used BF16 precision during training and inference. For detailed hyperparameter configurations, please refer to Table \ref{tab: exp_details}.
\begin{table}[ht]
\centering
\begin{tabular}{c|c}
\toprule
               & Hyperparameter \\
\midrule
Batch size     & $16$           \\
Learning rate  & $5e^{-5}$      \\
Warp-up steps   & $100$          \\
Training steps & $60k$          \\
Max grad norm  & $1.0$          \\
Precision      & BF16           \\
\bottomrule
\end{tabular}
\caption{Experimental settings for our TriFlow during the training stage.}
\label{tab: exp_details}
\end{table}

\begin{figure*}[ht]
\centering
\includegraphics[width=0.95\linewidth]{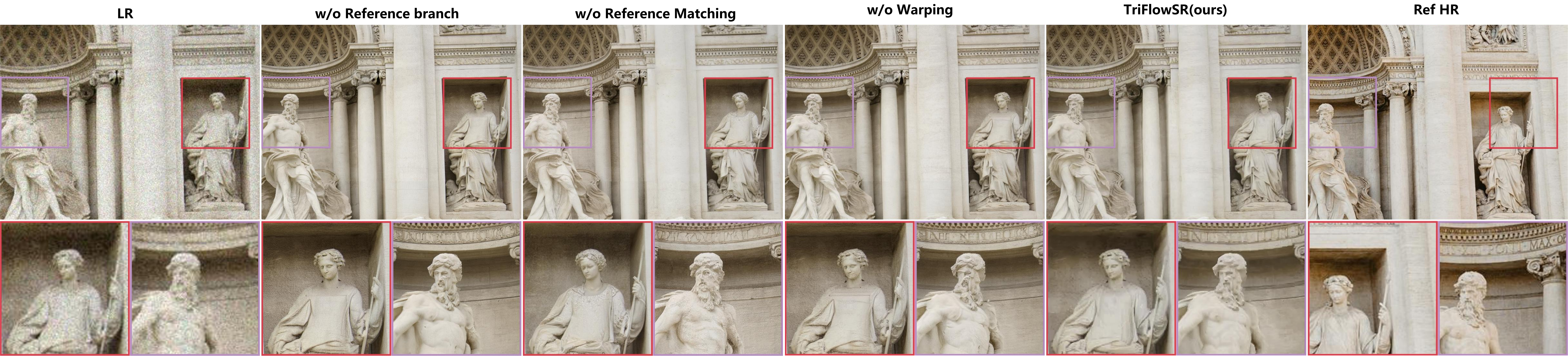} 
\caption{Visual comparisons of the Super-Resolution results of the ablation experiments on our proposed Lanmark-4K dataset. Without the Reference branch, ours will degenerate into a SISR model. If the Reference Matching Strategy \& Warping is not used, the LR image and the reference HR image may fail to align effectively, resulting in difficulty in matching some detailed information (such as fingers and face). Please zoom in for a better view.}
\label{fig: effective compares}
\end{figure*}
\subsection{Effectiveness of the Components}
\begin{table}[ht]\small
\centering
\begin{tabular}{ccc|cccc}
\toprule
RB & M & W & PSNR↑ & SSIM↑ & LPIPS↓ & DISTS↓\\
\midrule
           &              &         & 24.07 & 0.664 & 0.355  & 0.166  \\
\checkmark          &              &         &\underline{25.67}       &\underline{0.756}      &0.241        &\underline{0.113}        \\
\checkmark          & \checkmark            &         & 25.37 & 0.750  & \underline{0.239}  & \textbf{0.110}   \\
\checkmark          & \checkmark            & \checkmark       & \textbf{25.89} & \textbf{0.776} & \textbf{0.230}  & 0.115  \\
\bottomrule
\end{tabular}
\caption{Ablation study on our proposed Landmark-4K dataset. We report PSNR, SSIM, LPIPS and DISTS. RB represents the Reference branch, M represents the Matching operation and W represents the Warping operation. The best and second-best results are highlighted in \textbf{bold} and \underline{underlined}. “$\uparrow$" (resp. “$\downarrow$") means the larger (resp. smaller), the better.}
\label{tab: ablation_appendix}
\end{table}
Notably, the LR and Ref features are only concatenated during cross-attention operations, making them completely decoupled. First, we attempt to perform inference with only the LR branch and SR branch, which degenerates our model into a SISR model. Then, we introduce the Reference branch, but only utilize the relative position of the LR image tiles to map to the reference HR image tiles. Next, we use the pre-trained matching model to obtain the matching map and simply resize the corresponding reference region to the input tile size without using warping to align the images. Finally, we apply the warping operation to align the LR image with the reference HR image. As shown in Table \ref{tab: ablation}, compared to pure SISR, introducing the Reference branch can significantly improve PSNR, SSIM, LPIPS, and DISTS. Introducing the Reference Matching Strategy can solve the problem of incorrect tile retrieval due to scale mismatch and positional differences between the LR image and the reference HR image. As shown in Figure \ref{fig: effective compares}, without the Reference branch, the model is unable to utilize any information from the reference HR image, and the result is generated solely based on the diffusion prior. Due to the scale mismatch and positional differences between the LR image and the reference HR image, without using the Reference Matching Strategy, the LR image tile is likely to fail to correspond to the correct reference HR tile, resulting in incorrect reference information. When the Warping operation is not used, the LR image and the reference HR image may fail to align effectively, resulting in difficulty in matching some detailed information (such as fingers and face).
\begin{figure*}[ht]
\centering
\includegraphics[width=0.95\linewidth]{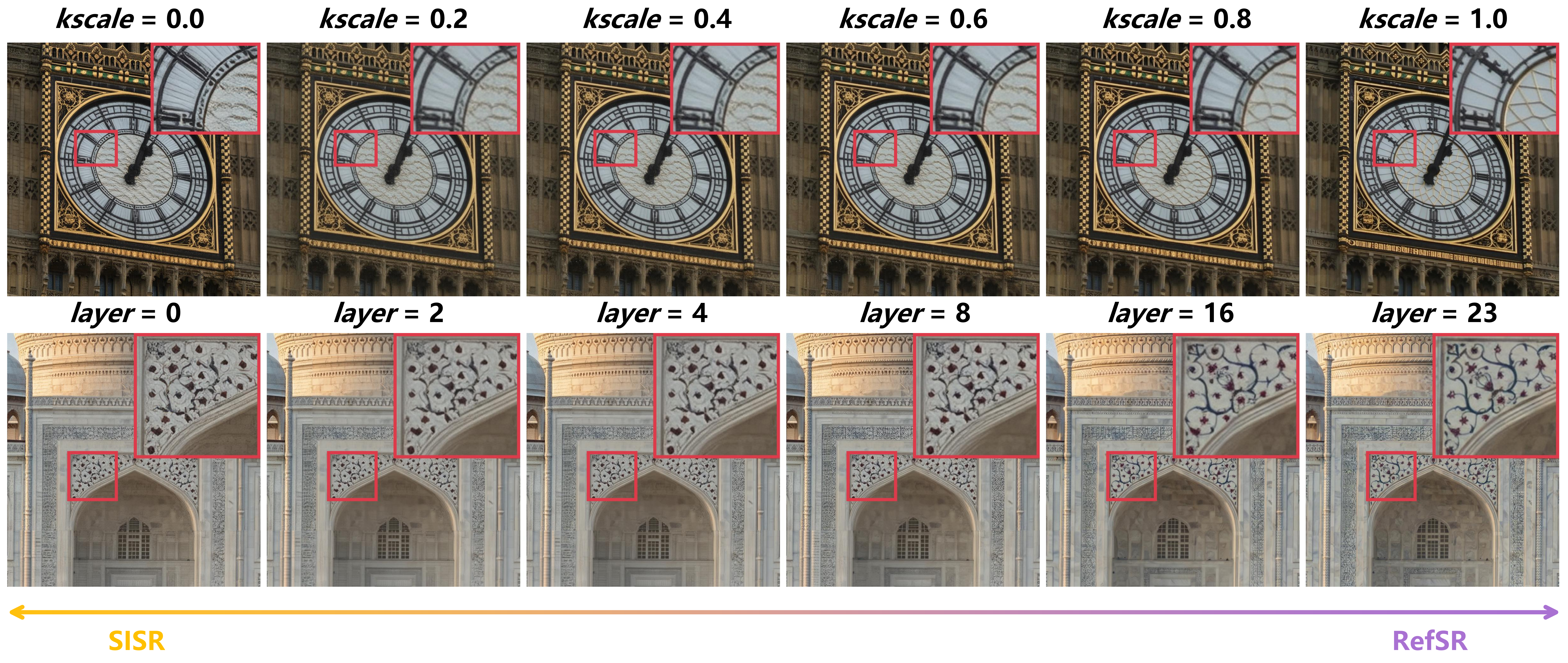} 
\caption{Visual comparisons of the Super-Resolution results of the ablation experiments on our proposed Lanmark-4K dataset. Without the Reference branch, ours will degenerate into a SISR model. If the Reference Matching Strategy \& Warping is not used, the LR image and the reference HR image may fail to align effectively, resulting in difficulty in matching some detailed information (such as facial information). Please zoom in for a better view.}
\label{fig: control compares}
\end{figure*}

\subsection{Control of the Reference branch}
Similar to ControlNet \cite{zhang2023adding} and IP-adapter \cite{ye2023ip}, TriFlow can also control the influence of the reference branch by adjusting the weight of the attention map. Specifically, during the Pat-Ref Attention operation, we can control the attention weight corresponding to the reference branch feature by multiplying $K^{Ref}$ by a coefficient $kscale$, i.e., $K^{Ref} \mapsto kscale \times K^{Ref}$. Since we only performed concatenation on the $K$ and $V$ features, directly controlling the weight of $K^{ref}$ can control the corresponding attention weight $Q [kscale \times K^{ref}]^{T}$ of $V^{ref}$, thereby achieving the result of controlling the influence of the reference branch. As shown in Table \ref{tab: ref_control}, when $kscale=1$, ours is a RefSR model, while when $kscale$ is between $0 - 1$, we can consider it as a trade-off between SISR and RefSR. As $kscale$ gradually decreases to zero, ours gradually degenerates into a SISR model. Moreover, since the LR branch and the Reference branch are completely decoupled, the degree of control over the reference HR image can also be effectively controlled by adjusting the number of layers in the Reference branch. As shown in Table \ref{tab: layer_control}, when the number of layers decreases (these models are all retrained in stage 2), our model gradually degenerates from a RefSR model to a SISR model, which is consistent with the principle of $kscale$. As shown in Figure \ref{fig: control compares}, we can control the degree of supervision from the reference HR image by adjusting the weight of the attention map ($k_{scale}$) and the number of layers in the Reference branch module, allowing our model to transition seamlessly between the SISR model and the RefSR model.

\begin{table}[ht]\small
\centering
\begin{tabular}{c|cccc}
\toprule
$kscale$ & PSNR↑ & SSIM↑ & LPIPS↓ & DISTS↓ \\
\midrule
0        & 24.07 & 0.664 & 0.355  & 0.166  \\
0.2      & 24.36 & 0.712 & 0.299  & 0.181  \\
0.4      & 24.57 & 0.709 & 0.287  & 0.159  \\
0.6      & 24.65 & 0.710 & 0.285  & 0.150  \\
0.8      & \underline{24.79} & \underline{0.716} & \underline{0.280}  & \underline{0.145} \\
1        & \textbf{25.89} & \textbf{0.776} & \textbf{0.230}  & \textbf{0.115}  \\
\bottomrule
\end{tabular}
\caption{Quantitative comparisons about the control scale ($kscale$) of the Reference branch on our proposed Landmark-4K dataset. We report PSNR, SSIM, LPIPS and DISTS. The best and second-best results are highlighted in \textbf{bold} and \underline{underlined}. “$\uparrow$" (resp. “$\downarrow$") means the larger (resp. smaller), the better.}
\label{tab: ref_control_appendix}
\end{table}

\begin{table}[ht]\small
\centering
\begin{tabular}{c|cccc}
\toprule
$layers$ & PSNR↑          & SSIM↑          & LPIPS↓         & DISTS↓         \\ 
\midrule
0        & 24.07          & 0.664          & 0.355          & 0.166          \\
2        & 24.74          & 0.723          & 0.28           & 0.132          \\
4        & 24.75          & 0.720           & 0.283          & 0.136         \\
8        & 24.79          & 0.724          & 0.273          & 0.132          \\
16       & \underline{25.38}          & \underline{0.747}          & \underline{0.252}          & \underline{0.126}          \\
23       & \textbf{25.89} & \textbf{0.776} & \textbf{0.230} & \textbf{0.115} \\ 
\bottomrule
\end{tabular}
\caption{Quantitative comparisons about the $layers$ of the Reference branch on our proposed Landmark-4K dataset. We report PSNR, SSIM, LPIPS, FID and DISTS. The best and second-best results are highlighted in \textbf{bold} and \underline{underlined}. “$\uparrow$" (resp. “$\downarrow$") means the larger (resp. smaller), the better.}
\label{tab: layer_control}
\end{table}

\subsection{Reference Image Condition Modeling}
In this chapter, we explored the current mainstream methods for introducing additional supervisory information into diffusion-based models, and compared them with our proposed Patch-Ref Attention mechanism.

\textbf{Controlnet} \cite{zhang2023adding} is an extension of diffusion models that enables pixel-level control over image generation using additional structural inputs, such as edge maps, depth maps, or human poses (for the RefSR task, it is the reference HR image). ControlNet introduces additional information by adding an auxiliary branch network that mirrors the structure of the backbone. A zero convolution module is used to integrate the noisy input with the conditional input within this branch. The output of the auxiliary branch is then added to the main network to guide the diffusion process. However, this approach lacks explicit pattern matching. The direct addition operation only allows for a coarse transfer of semantic information, making it difficult to effectively capture the structural and textural details of the reference high-resolution image. As shown in Table \ref{tab: ref_model}, we retrain the Reference branch based on the ControlNet structure using exactly the same settings and architecture. Experimental results demonstrate that our proposed TriFlowSR enables explicit feature matching between the LR image and the reference HR image, thereby making more effective use of the semantic and textural information in the reference HR image. 
\begin{table}[ht]\small
\centering
\begin{tabular}{c|cccc}
\toprule
$Method$ & PSNR↑ & SSIM↑ & LPIPS↓ & DISTS↓ \\
\midrule
Controlnet        & 24.55 & 0.717 & 0.271  & 0.121  \\
\textbf{TriFlowSR(ours)}        & \textbf{25.89} & \textbf{0.776} & \textbf{0.230}  & \textbf{0.115}  \\
\bottomrule
\end{tabular}
\caption{Quantitative comparisons about the reference image condition modeling of the Reference branch on our proposed Landmark-4K dataset. We report PSNR, SSIM, LPIPS and DISTS. The best results are highlighted in \textbf{bold}. “$\uparrow$" (resp. “$\downarrow$") means the larger (resp. smaller), the better.}
\label{tab: ref_model}
\end{table}

\textbf{IP-Adapter} \cite{ye2023ip} is a lightweight module designed to enable image-based prompting for diffusion models. It connects a pre-trained image encoder (e.g., CLIP) to the diffusion model, allowing users to guide image generation using reference images instead of text. By injecting features from the reference image into the cross-attention layers, IP-Adapter provides effective visual control while keeping the base model frozen. However, the introduced reference information is compressed by an image encoder, which typically retains only semantic and partial structural information. This injection method severely compromises the texture information of the reference HR image, leading to strong generative artifacts.

\textbf{T2I-Adapter} \cite{mou2024t2i} is a lightweight module designed to provide structural guidance (e.g., edges, depth, pose) to pre-trained text-to-image diffusion models. It works by learning an external adapter network that extracts features from the control input and injects them into the diffusion model without modifying its original weights. This approach enables controllable image generation with minimal computational cost and strong compatibility with existing models. The main difference between T2I-Adapter and IP-adapter is that the embedding information is introduced through addition (similar to ControlNet) rather than cross-attention. Like IP-adapter, T2I-Adapter also faces the issue of compressed reference information.


\subsection{Resource efficiency}
We compare the floating point of operations (FLOPs) and inference time with other methods on our proposed Lanmark-4K dataset by $torch.profiler$, using $1024 \times 1024$ images as the input. All experiments were conducted on a single NVIDIA H20. As shown in Table \ref{tab: resource_efficiency}, compared to other methods, our TriFlowSR has the lowest TFLOPs and the shortest inference time.
\begin{table}[ht]\small
\centering
\begin{tabular}{c|ccccc}
\toprule
Methods         & PSNR↑         & LPIPS↓        & TFLOPs↓ & time (s)↓ \\
\midrule
SUPIR           & 24.44          & 0.312         & \underline{1287.10} & \underline{17.25}    \\
CoSeR           & 24.72          & 0.359         & -       & 53.27    \\
ReFIR           & \underline{25.21}         & \underline{0.28}          & 1561.51 & 26.07    \\
TriFlowSR(ours) & \textbf{25.89}& \textbf{0.23} & \textbf{491.65}  & \textbf{10.59}   \\
\bottomrule
\end{tabular}
\caption{Quantitative comparisons about resource efficiency on our proposed Landmark-4K dataset. We report PSNR, LPIPS, TFLOPs and inference time. The best and second-best results are highlighted in \textbf{bold} and \underline{underlined}. “$\uparrow$" (resp. “$\downarrow$") means the larger (resp. smaller), the better.}
\label{tab: resource_efficiency}
\end{table}

\subsection{More Visual Results}
\begin{figure*}[ht]
\centering
\includegraphics[width=0.9\textwidth]{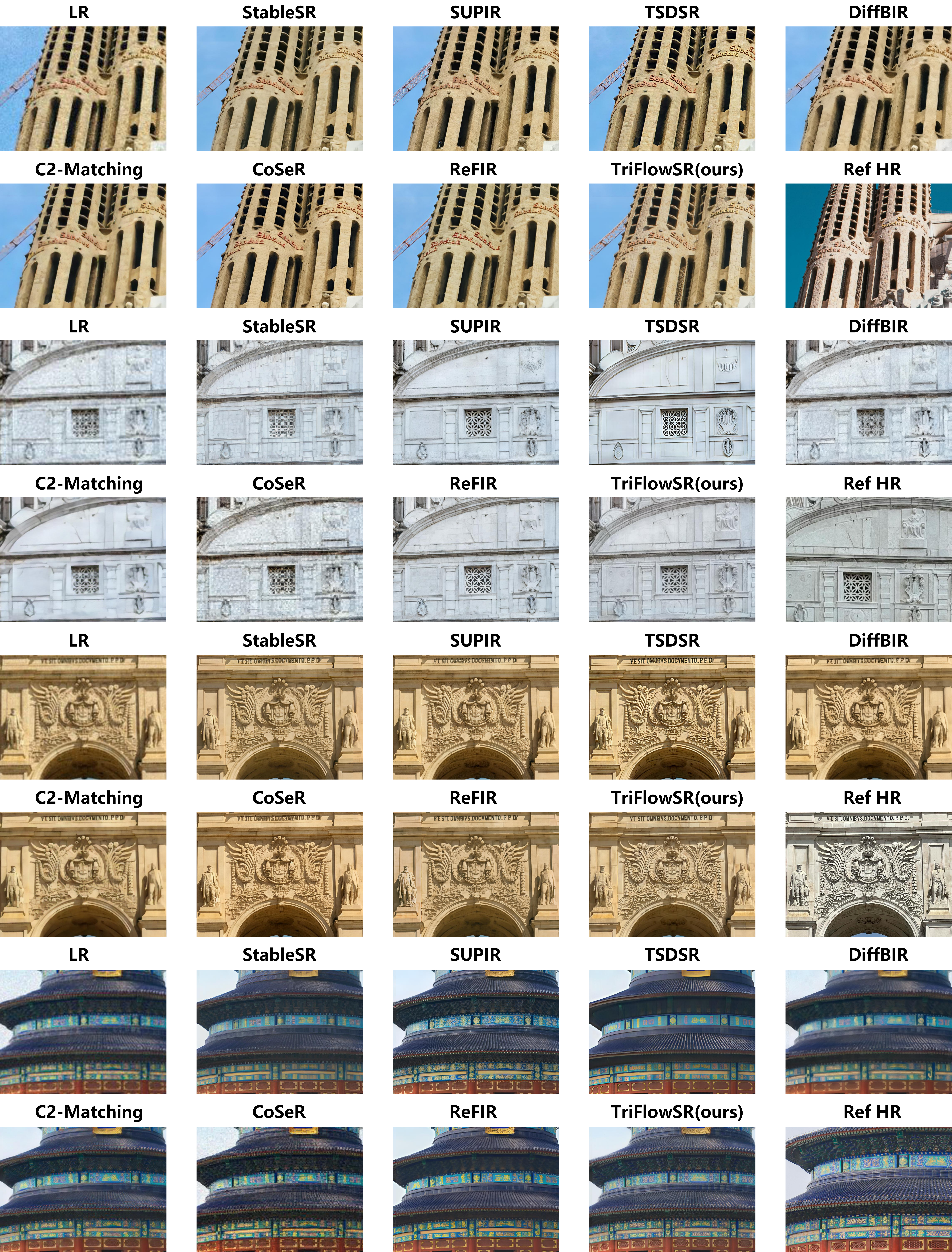} 
\caption{More visual comparisons of the Super-Resolution results by different methods on the our proposed Lanmark-4K dataset. Please zoom in for a better view.}
\label{fig: landmark-compares-appendix}
\end{figure*}

\begin{figure*}[ht]
\centering
\includegraphics[width=0.95\textwidth]{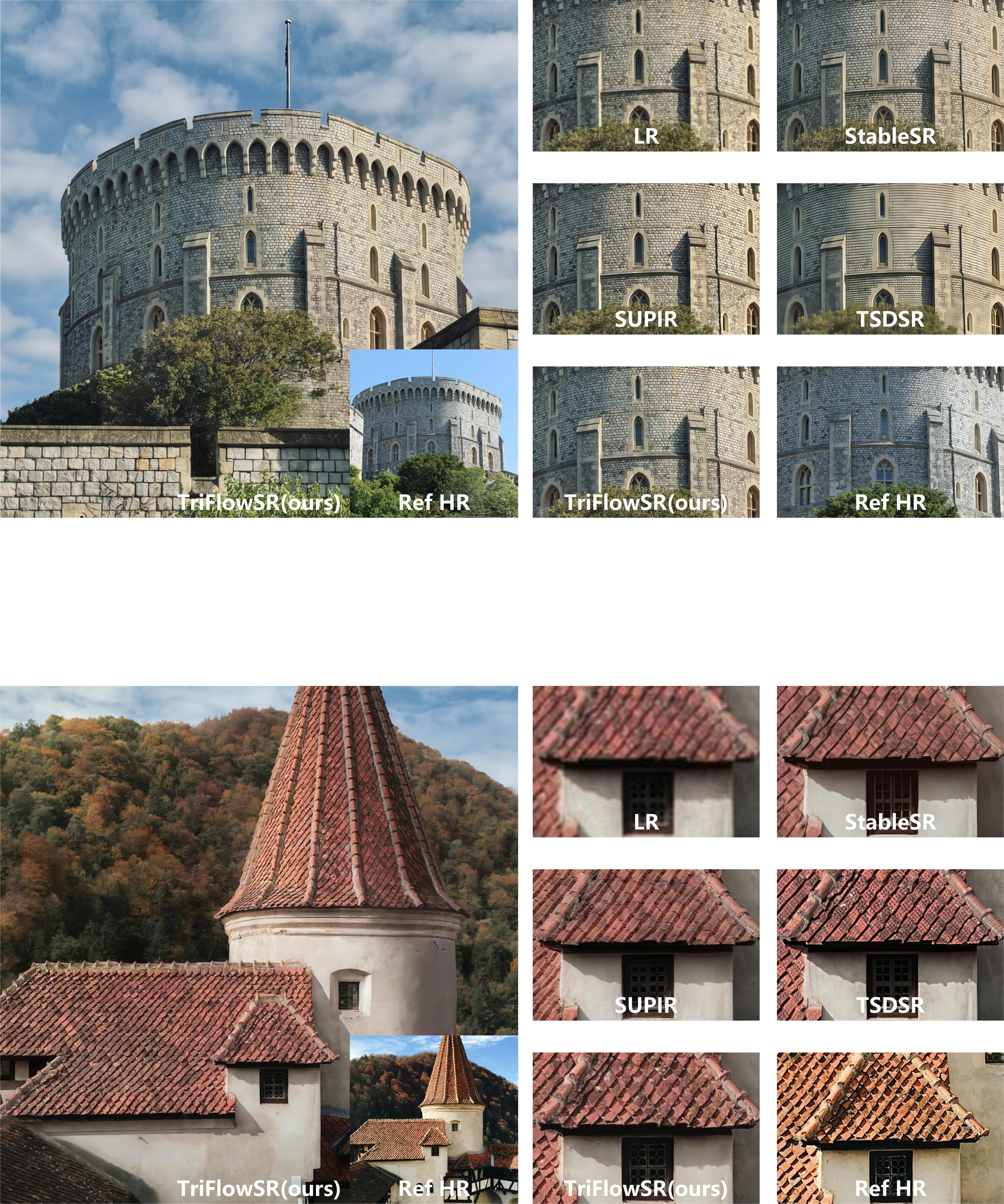} 
\caption{More visual comparisons of the Super-Resolution results by different methods on the our proposed Lanmark-4K dataset. Please zoom in for a better view.}
\label{fig: landmark-compares-appendix-1}
\end{figure*}

\begin{figure*}[ht]
\centering
\includegraphics[width=0.95\textwidth]{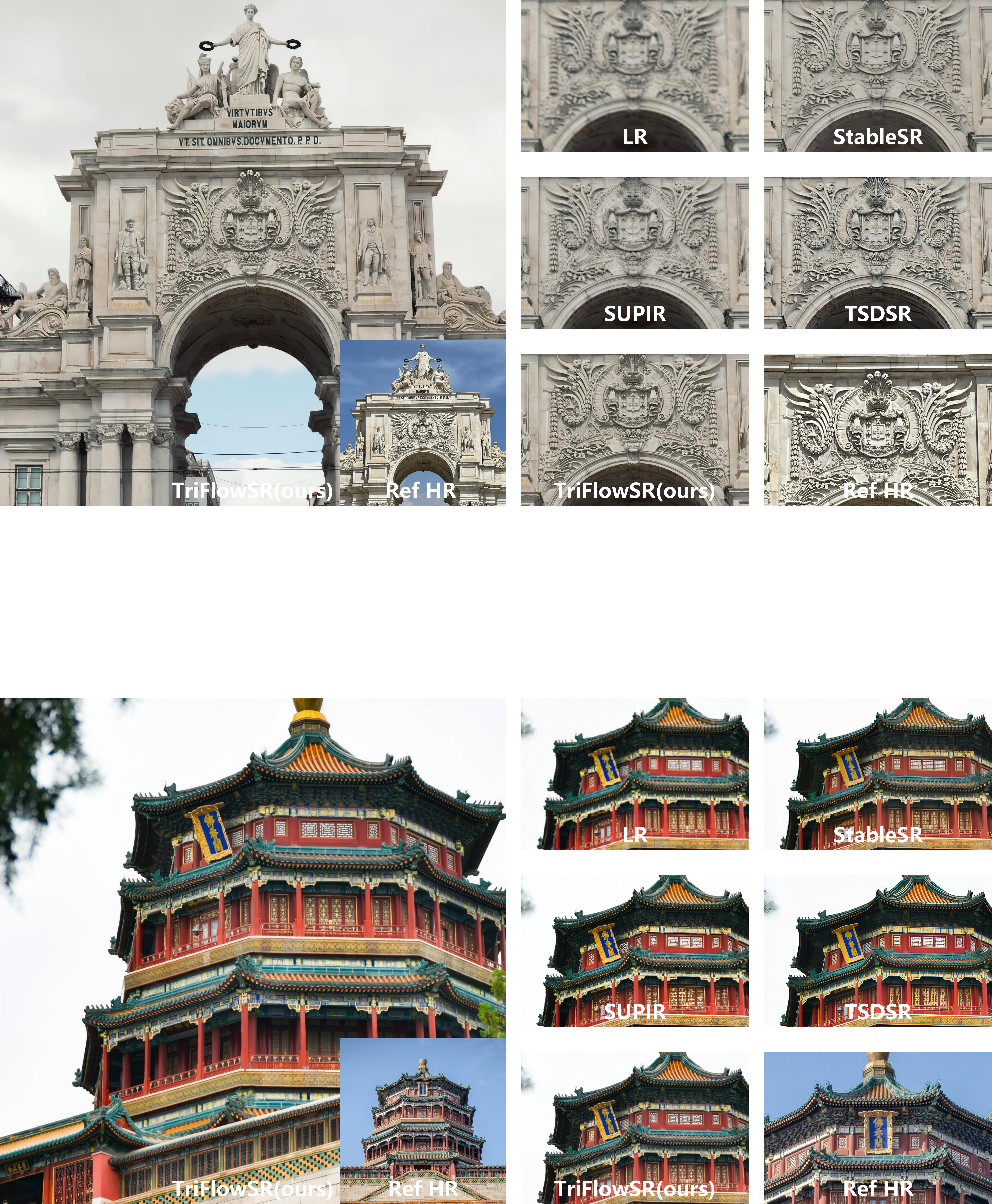} 
\caption{More visual comparisons of the Super-Resolution results by different methods on the our proposed Lanmark-4K dataset. Please zoom in for a better view.}
\label{fig: landmark-compares-appendix-2}
\end{figure*}

\begin{figure*}[ht]
\centering
\includegraphics[width=0.95\textwidth]{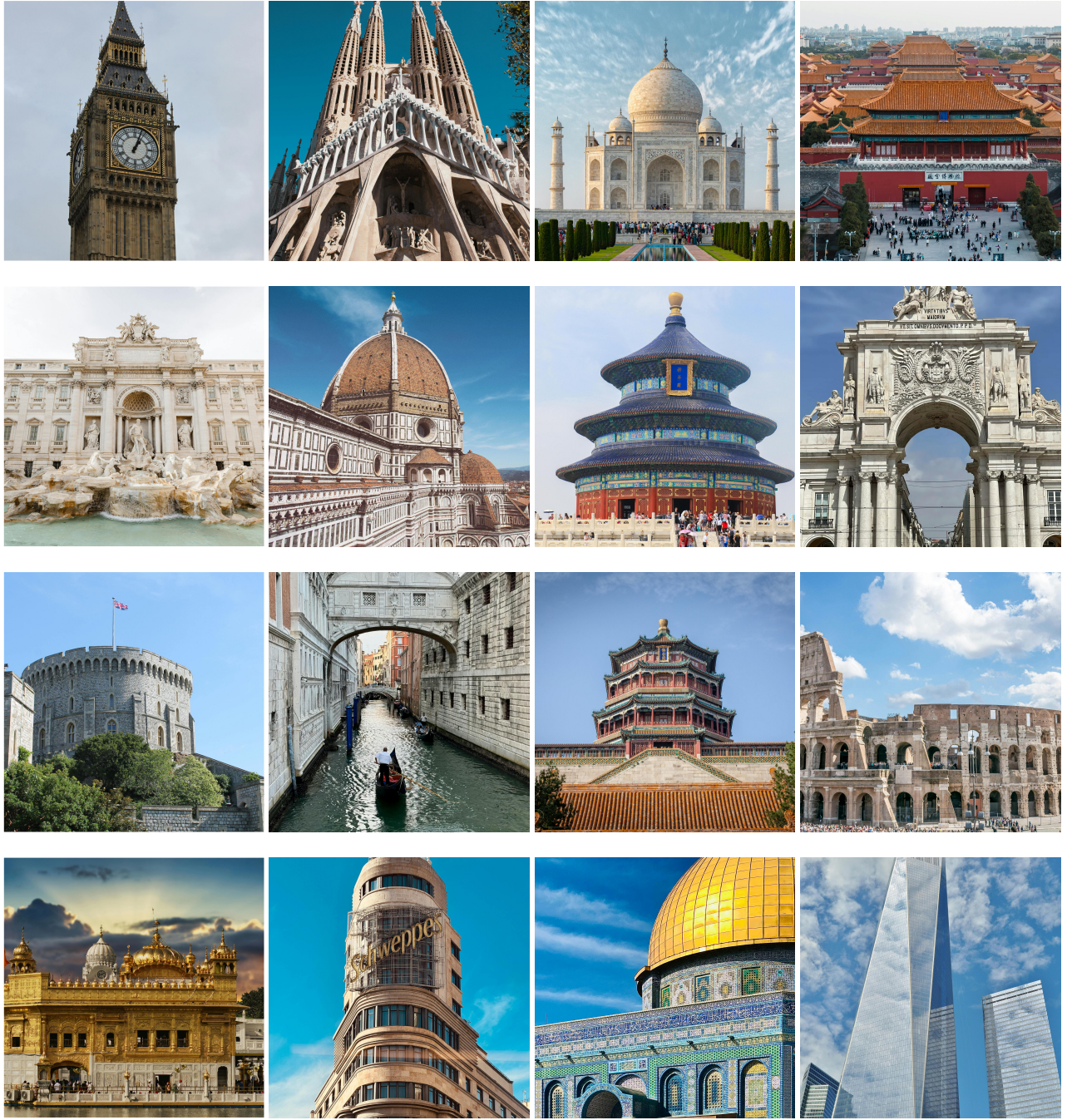} 
\caption{More visual demonstrations on our proposed Landmark-4K dataset. Please zoom in for a better view.}
\label{fig: landmark-show-appendix}
\end{figure*}

As shown in the Figure \ref{fig: landmark-compares-appendix}, Figure \ref{fig: landmark-compares-appendix-1} and Figure \ref{fig: landmark-compares-appendix-2}, we provide more visual comparison results. Meanwhile, we also provide more illustrations about our proposed Landmark-4K dataset, as shown in the Figure \ref{fig: landmark-show-appendix}.


\end{document}